\documentclass{article}

\usepackage{preprint,times}

\usepackage[T1]{fontenc}
\usepackage[utf8]{inputenc}
\usepackage{microtype}
\usepackage{inconsolata}

\usepackage{amsmath}
\usepackage{amssymb}
\usepackage{hyperref}
\usepackage{enumitem}
\usepackage{url}
\usepackage{booktabs}
\usepackage{graphicx}
\usepackage{listings}
\usepackage{xcolor}
\usepackage{multirow}
\usepackage{makecell}
\usepackage{wrapfig}

\definecolor{darkblue}{rgb}{0, 0, 0.5}
\hypersetup{colorlinks=true, citecolor=darkblue, linkcolor=darkblue, urlcolor=darkblue}

\title{\textit{Vision Wormhole}: Latent-Space Communication\\ in Heterogeneous Multi-Agent Systems}

\author{
\textbf{Xiaoze Liu\textsuperscript{1,*}, Ruowang Zhang\textsuperscript{1,2,*}, Weichen Yu\textsuperscript{3},}\\
\textbf{Siheng Xiong\textsuperscript{4}, Liu He\textsuperscript{1}, Feijie Wu\textsuperscript{1}, Hoin Jung\textsuperscript{1},}\\
\textbf{Matt Fredrikson\textsuperscript{3}, Xiaoqian Wang\textsuperscript{1}, Jing Gao\textsuperscript{1}}\\
{\normalfont\small \textsuperscript{1}Purdue University \quad \textsuperscript{2}Contextual AI}\\
{\normalfont\small \textsuperscript{3}Carnegie Mellon University \quad \textsuperscript{4}Georgia Institute of Technology}\\
{\normalfont\small \textsuperscript{*}Equal contribution.}\\
{\normalfont\footnotesize \texttt{\{xiaoze,zhan5763,he425,wu1977,jung414,joywang,jinggao\}@purdue.edu}}\\
{\normalfont\footnotesize \texttt{\{weichenyu,mfredrik\}@cmu.edu} \quad \texttt{sxiong45@gatech.edu}}
}

\hypersetup{
    pdftitle={Vision Wormhole: Latent-Space Communication in Heterogeneous Multi-Agent Systems},
    pdfauthor={Xiaoze Liu, Ruowang Zhang, Weichen Yu, Siheng Xiong, Liu He, Feijie Wu, Hoin Jung, Matt Fredrikson, Xiaoqian Wang, Jing Gao},
    pdfsubject={Latent-space communication in heterogeneous multi-agent systems},
    pdfkeywords={latent communication, multi-agent systems, vision-language models, model composition}
}
\begin{document}

\raggedbottom
\maketitle

\begin{abstract}
Heterogeneous multi-agent systems combine models with different capabilities through a common communication interface. Exchanging internal states directly requires translating between model-specific representations and controlling intermediate computation. We introduce the \textbf{Vision Wormhole}, which repurposes the visual input interface of Vision-Language Models (VLMs) for continuous communication between frozen heterogeneous agents. A Universal Visual Codec encodes each sender's latent rollout into a fixed-size message, maps it through a shared reference space, and decodes received messages into the receiver's image-token span. Per-model codecs and affine reference maps form a hub-and-spoke architecture with $O(N)$ components for $N$ models. Each model learns its codec independently through self-distillation on anchor texts, and shared-anchor alignment enables reuse across communication partners. Across four VLM families, six team configurations, and nine reasoning benchmarks, Vision Wormhole improves accuracy by 6.0 percentage points on average over text-mediated MAS and achieves a 1.69$\times$ geometric-mean speedup in batch-normalized end-to-end runtime.
\end{abstract}

\begin{figure}[!h]
    \centering
    \includegraphics[width=.8\textwidth]{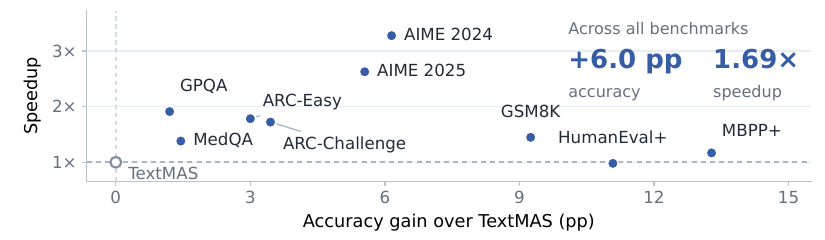}

    \caption{\textbf{Joint accuracy and speed gains.} Each point averages all six configurations for one benchmark; speedup uses the geometric mean of TextMAS/VW runtime ratios.}
    \label{fig:teaser}

\end{figure}

\section{Introduction}
\label{sec:intro}

Multi-agent systems (MAS) offer a way to bring the capabilities of several pretrained models into a single problem-solving process. Agents plan, critique, and refine each other's work, making the output of one stage the context for the next \citep{guo2024large,wu2024autogen,hong2023metagpt,li2023camel,zhang2024chain}. Model diversity gives this composition practical value: different models excel at different agent functions and task domains, and assigning heterogeneous models to the same workflow improves performance \citep{ye2025xmasbuildingmultiagentsystems}. Gains from aggregating diverse model responses and heterogeneous debate further demonstrate the value of combining model families \citep{wang2024mixtureofagentsenhanceslargelanguage,zhang2025stopovervaluingmultiagentdebate}. The opportunity is to make each model's intermediate computation available to the rest of the team.

Communication determines how that computation reaches another model. In text-mediated teams, each handoff generates a response token by token, which subsequent agents process as additional context \citep{wu2024autogen,wang2024mixtureofagentsenhanceslargelanguage}. Repeating this cycle across roles adds generation and context-processing costs. Latent communication exposes a direct route through hidden activations and attention caches \citep{ramesh2025activations,zou2025latent,fu2025cache}. For heterogeneous teams, embedding transformations and cache translation provide ways to connect independently pretrained models \citep{du-etal-2026-enabling,fu2025cache}. We build on this direction with a concrete goal: \emph{let a frozen model use another model's internal states as context through an interface that can be reused across model families.} Realizing this goal brings together three challenges.

\begin{figure}[!t]
    \centering
    \includegraphics[width=\textwidth]{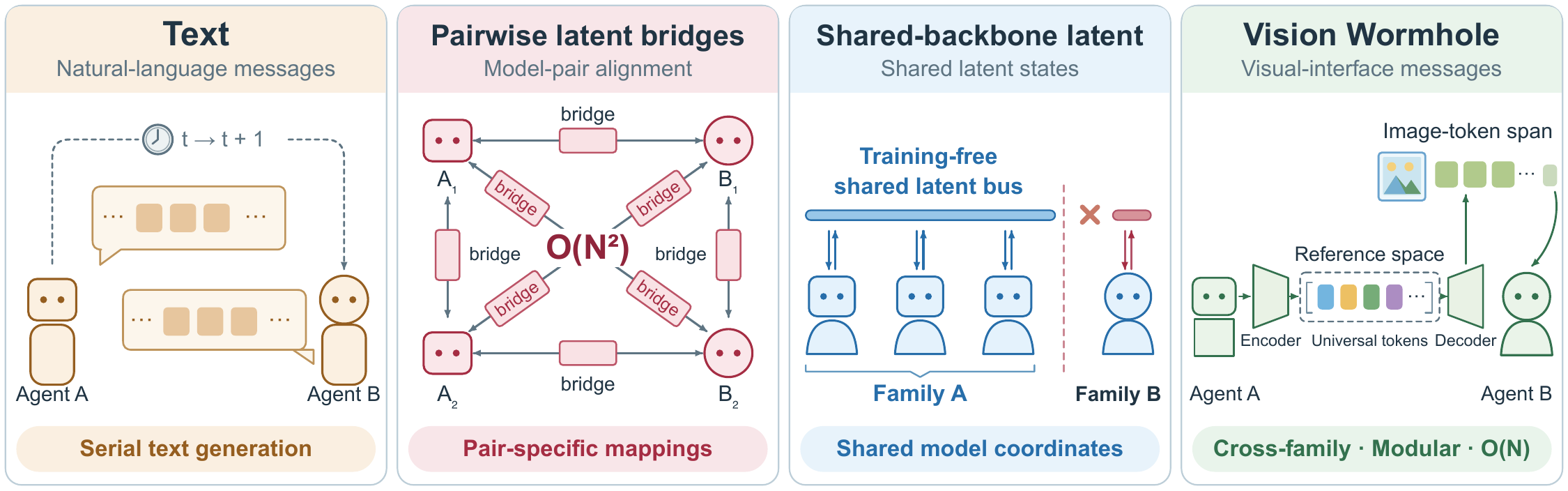}
    \caption{\textbf{The Vision Wormhole.} The visual input interface provides a destination for continuous messages between heterogeneous VLM agents. Per-model codecs connect the sender and receiver through a shared reference space.}
    \label{fig:intro_demo}

\end{figure}

\noindent\textbf{Representation compatibility.}
An internal state acquires its meaning through the model that produced it. Across model families, its width, feature coordinates, and role in the computation can differ. Cache translation addresses layer and key--value layouts \citep{fu2025cache,chen2026iseeknowi}. Usable communication therefore requires adapting a message to both the structure and the conditioning behavior of its receiver. The choice of where a message enters the receiver is central to this adaptation.

\noindent\textbf{Scalable model integration.}
Learned bridges \citep{fu2025cache,du-etal-2026-enabling,yang2026xbridgeentitygroundedlatentbridge} usually train mappings for specific sender--receiver pairs. Supporting every directed pair in a pool of $N$ models then entails $N(N-1)$ bridges, tying adaptation work to the set of communication partners. Shared-space approaches make model interfaces reusable through per-model mappings and interaction modules \citep{fein2025mixture}. For heterogeneous teams, this modularity is valuable because a model can serve several roles and communicate with several partners. The challenge is to retain receiver-specific adaptation within a common message format.

\noindent\textbf{Supervision for communication.}
Continuous messages need a learning signal that connects their contents to the receiver's behavior. Existing approaches obtain this signal from receiver-side reconstruction and generation objectives \citep{chen2026iseeknowi} or optimize multi-agent interaction trajectories \citep{yu2026learningcommunicateendtoendoptimization}. A reusable interface has two learning requirements: each module must learn to encode and consume context within its own frozen model, and the message spaces learned by separate modules must support exchange across models. Meeting both requirements calls for local conditioning targets and a common source of cross-model correspondences.

We propose the \textbf{Vision Wormhole} (VW), which turns the visual input interface of vision-language models (VLMs) into a communication channel between heterogeneous agents (Figure~\ref{fig:intro_demo}). The key insight is that multimodal pretraining already equips a language backbone to read continuous features supplied by a visual encoder \citep{alayrac2022flamingo,li2023blip2}. Visual conditioning also supports language-rich content, including rendered reasoning traces, documents, and code \citep{wang2026renderofthoughtrenderingtextualchainofthought,wei2025deepseekocrcontextsopticalcompression,shi2026codeocreffectivenessvisionlanguage}. We repurpose this capability for model-to-model communication: \emph{a learned codec presents one model's internal states as visual context to another frozen model.} The visual pathway supplies a pretrained receiver interface; the codec learns how to express a latent message through it.

Each backbone receives a \emph{Universal Visual Codec} with an encoder and a decoder. An intermediate agent produces a latent rollout by feeding its hidden states back as continuous inputs. The encoder converts this rollout into fixed-size message tokens, and the receiver's decoder expresses incoming messages in its image-token span. This context conditions the frozen receiver's subsequent computation. Intermediate agents exchange these continuous messages, and the final agent generates the answer in text.

To make the codecs reusable across partners, we connect their message spaces through a shared reference space. Per-model affine maps translate messages into this space for exchange and aggregation, then back into a receiver's codec coordinates for decoding. The resulting hub-and-spoke architecture has $O(N)$ components: one codec per model and two maps per non-reference model. Each backbone's visual decoder serves all of its communication partners.

We train this interface in two stages using anchor texts. First, a frozen backbone reads an anchor as text and supplies its prompt-boundary state and next-token distribution as self-distillation targets. Its codec encodes the associated latent rollout and decodes it into visual context. We train the codec so that the same backbone, conditioned on this visual context, reproduces the text-conditioned targets. Each model thereby supplies the supervision for its own encoder and decoder. Second, we give every model the same anchor texts and encode the resulting rollouts to obtain corresponding messages. We fit the affine reference maps to these messages by ridge regression. This separates local codec training from fitting the cross-model correspondences, allowing independently trained codecs to be composed through the reference space. The backbone parameters remain fixed throughout both stages.

Across four VLM families, six team configurations, and nine benchmarks, VW improves accuracy by \textbf{6.0 percentage points on average} over text-mediated MAS and achieves a \textbf{1.69$\times$ geometric-mean speedup} in batch-normalized end-to-end runtime (Figure~\ref{fig:teaser}). Both summaries weight all 54 configuration--task pairs equally. The evaluation spans mathematical and scientific reasoning, commonsense reasoning, and code generation, using the same model assignments and role sequence for the two communication channels. These results establish the visual interface as a practical route to composing frozen heterogeneous models through continuous messages.

\paragraph{Contributions.}
Our contributions are fourfold:
\begin{itemize}[leftmargin=*, nosep]
    \item \textbf{Visual conditioning for latent communication.} We introduce a codec that makes sender-side latent rollouts available as continuous visual context to a frozen heterogeneous receiver.
    \item \textbf{Reusable interfaces with $O(N)$ components.} We compose per-model codecs through affine reference maps, allowing the same encoder and decoder to serve multiple communication partners.
    \item \textbf{Independent codec training from text.} We develop a two-stage recipe that uses local self-distillation to learn visual conditioning and shared anchor texts to align the message spaces.
    \item \textbf{Joint accuracy and runtime gains.} Across four VLM families and nine benchmarks, VW improves accuracy by 6.0 percentage points on average and achieves a 1.69$\times$ overall speedup over text-mediated MAS.
\end{itemize}

\section{Related Work}
\label{sec:related_work_main}

\paragraph{Multi-agent coordination.}
LLM-based MAS combine role specialization, planning, memory, and tool use through orchestration frameworks and collaborative workflows \citep{guo2024large,wu2024autogen,hong2023metagpt,li2023camel,zhang2024chain,zhao2025connecting,zhou2025reso,zhang2025survey}. Model selection adds another design axis: function--domain evaluations guide heterogeneous role assignments, and layered response aggregation composes several models' outputs \citep{ye2025xmasbuildingmultiagentsystems,wang2024mixtureofagentsenhanceslargelanguage}. Memory modules retain information across interactions, and workflow optimization allocates computation among successive agent calls \citep{packer2023memgpt,zhong2024memorybank,hu2025hiagent,chen2025optima,zhuge2024language}. Communication design specifies the participants, message content, timing, and purpose of each exchange \citep{chen2026wsmultiagentcommunicationtalks}. Analyses of agent systems study information sharing, coordination, and specialization as factors in task performance \citep{pezeshkpour2024reasoning,cemri2025multi}. Our main comparison evaluates text and visual communication within the same workflow and model assignments.

\paragraph{Latent communication and reasoning.}
Continuous channels expose several forms of intermediate computation, with corresponding choices of alignment and receiver-side fusion \citep{liu2026tokensunifiedframeworklatent}. Probability-weighted vocabulary embeddings encode next-token distributions \citep{pham2024cipher}; activation and state-trajectory interfaces communicate residual-stream vectors or token-wise changes in hidden states \citep{ramesh2025activations,tang2025augmentingmultiagentcommunicationstate}. Cache-based collaboration exchanges attention state across agents, with selective layer sharing and reusable latent operations for planning, review, and voting \citep{zou2025latent,shi2026kvcommenablingefficientllm,jin2026agentprimitivesreusablelatent}. Execution-oriented approaches reuse caches across contexts and selectively recompute positions affected by prefix changes \citep{ye2025kvcomm,geng2026relaycachingacceleratingllmcollaboration}. Related work also studies message compression and efficient transport for latent communication \citep{kriuk2025qkvcommefficientmultiagentcommunication,li2026latentleadsbetterrelay,lee2026lowlatencyedgellmhandover,dai2026tokenkvcachecommunicationmediaselection}.

Heterogeneous communication learns transformations into a receiver's input embeddings or attention cache \citep{du-etal-2026-enabling,yang2025directsemanticcommunicationlarge,fu2025cache,rossi2026latentcacheflowmodeltomodel,chen2026iseeknowi}. Shared representation spaces support composition through per-model mappings, cross-attention, or routed expert interactions \citep{bansal2024llmaugmentedllmsexpanding,fein2025mixture,zheng2025thought}. Training objectives connect these interfaces to receiver behavior through cache reconstruction, generation supervision, or joint optimization over multi-agent trajectories \citep{chen2026iseeknowi,yu2026learningcommunicateendtoendoptimization}. Continuous reasoning develops recurrent computation within a model \citep{hao2024training,liu2024deliberation,qu2025survey}; internalized collaboration distills debate or composes latent role representations for a single model's inference \citep{yi2026latentagentsposttrainingprocedure,zeng2026modelmindsunlockingmultiagent}. The Vision Wormhole connects separately trained visual codecs through a shared message space. Appendix~\ref{app:latent_channels} specifies our heterogeneous baseline implementations.

\paragraph{Continuous conditioning and representation alignment.}
Learned soft prompts and prefixes condition frozen language models through trainable continuous inputs or layer-wise prefix states \citep{lester2021prompttuning,li2021prefixtuning}. Multimodal architectures use projectors and resamplers to connect visual features to language generation \citep{alayrac2022flamingo,li2023blip2}. These architectures establish visual features as a source of continuous context for a language backbone. Cross-modal representation learning also uses paired image--text data to align semantic information across modalities \citep{radford2021clip,jia2021align,zhai2023siglip}.

Across independently trained networks, representation alignment, model stitching, and permutation-aware merging study correspondences between internal feature spaces \citep{bansal2021revisiting,lenc2015equivariance,ainsworth2022git,wortsman2022model}. Anchor-based relative representations support reuse of neural components across spaces and modalities \citep{moschella2024latentcommunicationartificialneural}. Our visual codecs combine learned continuous conditioning with affine alignment on shared text anchors, providing a reusable communication interface for each frozen backbone. Appendix~\ref{app:related_work} discusses communication methods, model composition, compression, and channel analysis in greater depth.

\section{Method: The Vision Wormhole}
\label{sec:method}

\begin{figure}[t]
    \centering
    \includegraphics[width=\textwidth]{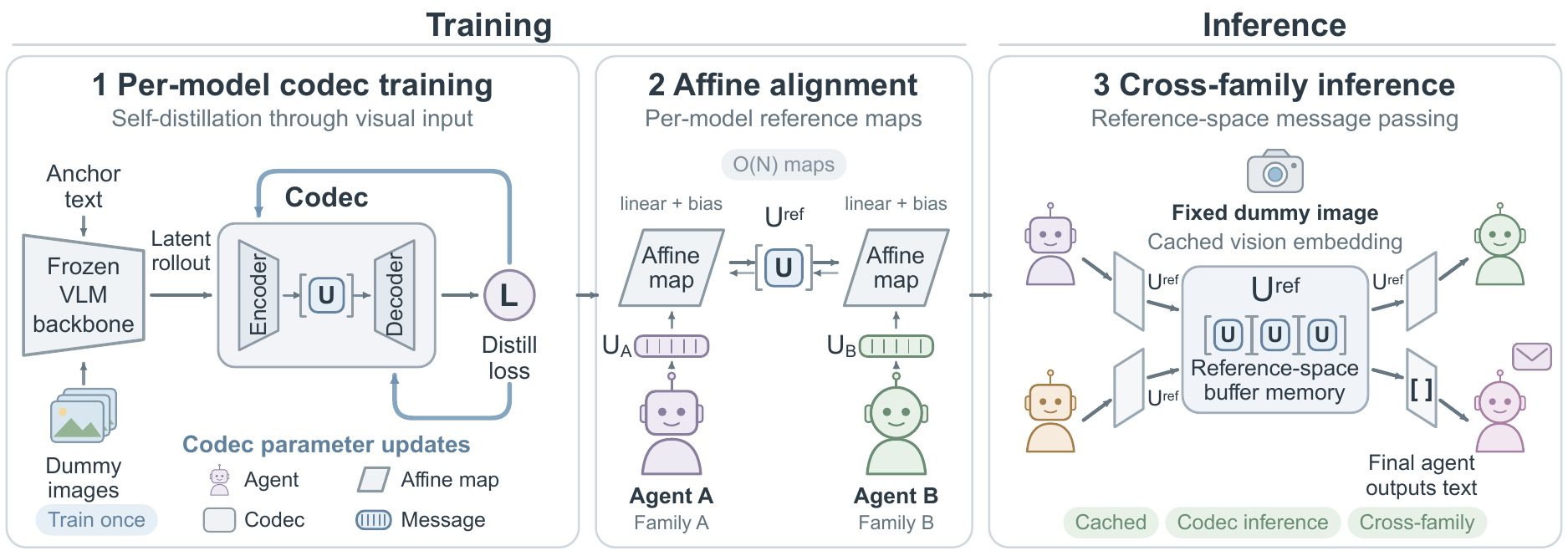}
    \caption{\textbf{Vision Wormhole overview.} A sender encodes its latent rollout into universal tokens and maps them to a shared reference space. The receiver maps incoming tokens to its codec coordinates and decodes them into a gated residual on its image-token span.}
    \label{fig:vision_wormhole_overview}

\end{figure}

The \textbf{Vision Wormhole} connects heterogeneous agents by writing continuous messages into the \emph{image-token span} of a VLM. Each backbone has a \emph{Universal Visual Codec} with an encoder and a decoder. The encoder maps a latent rollout to fixed-size universal tokens; affine maps connect these tokens to a shared reference space $\mathcal{U}$; and the receiver's decoder generates a gated perturbation of its image-token embeddings. Codecs are trained separately for each model and frozen at inference, and all VLM backbone parameters remain frozen throughout. Figure~\ref{fig:vision_wormhole_overview} illustrates the pipeline. Appendix~\ref{app:details_codec} provides architectural and procedural details.

\subsection{Preliminaries}
\label{sec:preliminaries}

\noindent\textbf{Multi-agent system (MAS).}
We define a multi-agent system as $\mathcal{S}=(\mathcal{A},\pi)$, where
$\mathcal{A}=\{a_1,\dots,a_N\}$ is a set of agents and $\pi$ is the orchestration policy.
Given an input query $q$, $\pi$ specifies role execution, interaction order, and routing among agents, and the system outputs a final answer $a$.
This definition is modality-agnostic, in that agents may communicate through discrete text (TextMAS) or through any other interface they expose.
In our experiments, each agent $a_i$ is a vision-language model (VLM) with frozen backbone $F_i$ and input embedding dimension $d_i$.
We focus on heterogeneous VLM-based MAS, where agent backbones may come from different VLM families.
VLM agents expose a continuous conditioning interface through \emph{visual token embeddings}, which serve as the destination for decoded messages.

\noindent\textbf{The VLM visual interface as a continuous channel.}
The VLMs in our experiments form an input embedding sequence by concatenating
(1) text token embeddings and
(2) a dedicated \emph{image-token span}.
Given an image $\mathbf{I}$, a vision encoder $E_{\textrm{vis}}^{(i)}$ and projector $\phi_i$ produce
$
X^{(i)}_{\mathrm{img}} = \phi_i(E^{(i)}_{\mathrm{vis}}(\mathbf{I})) \in \mathbb{R}^{L^{(i)}_{\mathrm{img}} \times d_i},
$
which is inserted into the language stream at model-specific image positions.
During multimodal training, $F_i$ learns to condition its language outputs on these projected image features. Our codec learns a continuous prompt at the same input positions.

\noindent\textbf{Latent rollouts.}
Let a VLM backbone $F_i$ process a prompt and produce the final hidden vector at the prompt boundary, $h^{(i)}_0 \in \mathbb{R}^{d_i}$.
We define a \emph{latent rollout} by repeatedly feeding back a single continuous pseudo-token embedding derived from the previous hidden state while reusing the prompt's attention cache (detailed in Appendix~\ref{app:details_rollout_cached}).
At step $t$, we form an input embedding
$
x^{(i)}_t = \mathrm{NormMatch}_i(h^{(i)}_t),
$
where $\mathrm{NormMatch}_i$ rescales vectors to match the typical norm of the model's token embeddings (see Eq.~\ref{eq:normmatch} in Appendix~\ref{app:details_codec_train}).
A $T$-length rollout yields
$
H_i = [x^{(i)}_0,\dots,x^{(i)}_{T-1}] \in \mathbb{R}^{T \times d_i},
$
which is the input to the sender's codec. Appendix~\ref{app:codec_training_setup} gives the codec training hyperparameters.

\noindent\textbf{Notation.}
For agent $i$, let $d_i$ denote its embedding dimension.
We use a universal token dimension $D$ shared across all agents.
A message is represented by $K_{\mathrm{u}} = K+2$ universal tokens (with $K$ message tokens plus two special tokens: a global token and a style token).
For vision-span writing, we decode to $K_{\mathrm{img}}$ \emph{image query tokens}.
We denote by $L^{(i)}_{\mathrm{img}}$ the number of image tokens used by agent $i$ in its prompt.

\subsection{Training a Vision Codec for a Specific VLM}
\label{sec:method:codec_train}

We train each codec through self-distillation on anchor texts. The teacher provides a prompt-boundary hidden state and next-token distribution from a text-conditioned forward pass. The student learns to reproduce these two quantities through vision-span conditioning (Appendix~\ref{app:details_distill_boundary}).

\noindent\textbf{(1) Sender message extraction via latent rollout.}
During training, the backbone processes a teacher prompt containing an anchor text $m$. Its boundary state $h^{(i)}_0$ initializes the length-$T_{\mathrm{train}}$ rollout defined in Section~\ref{sec:preliminaries}. At MAS inference, the rollout starts from the current role's task prompt and received context, as specified in Section~\ref{sec:method:inference}.

\noindent\textbf{(2) Universal-token encoder.}
We encode $H_i$ into a fixed-size set of universal tokens using a Perceiver resampler \citep{alayrac2022flamingo}, in which a fixed set of learned queries cross-attends to the rollout (Appendix~\ref{app:details_resampler}).
The encoder $\mathcal{E}_i$ outputs
$
U_i = \mathcal{E}_i(H_i) \in \mathbb{R}^{K_{\mathrm{u}} \times D}.
$
The encoder uses $K$ learned message queries and two special queries for global aggregation and style/statistics cues (Eq.~\ref{eq:style_stats} in Appendix~\ref{app:details_global_style}).

\noindent\textbf{(3) Universal-to-vision decoder and gated injection.}
The decoder $\mathcal{D}_i$ maps universal tokens to a vision-span perturbation and a scalar gate:
$
(\Delta_i, g_i) = \mathcal{D}_i(U_i), \qquad
\Delta_i \in \mathbb{R}^{K_{\mathrm{img}} \times d_i},\;\; g_i \in (0,1).
$
The perturbation $\Delta_i$ is a \emph{continuous prompt} in the agent's embedding space, and the gate $g_i$ scales its strength per example.
We parameterize the injected embeddings as a residual on a fixed baseline visual embedding.
Let $\bar{X}^{(i)}_{\mathrm{img}} \in \mathbb{R}^{L^{(i)}_{\mathrm{img}} \times d_i}$ denote the image-token embeddings induced by a fixed dummy image under the frozen VLM.
We resample $\Delta_i$ to the required image-span length via a length-resampling operator $\mathrm{Resample}(\cdot;L)$ and write
\begin{equation}
X^{(i)}_{\mathrm{img}} \;=\; \bar{X}^{(i)}_{\mathrm{img}} \;+\; g_i \cdot \mathrm{Resample}(\Delta_i; L^{(i)}_{\mathrm{img}}).
\label{eq:vision_inject}
\end{equation}
All downstream computation is performed by the frozen backbone conditioned on $X^{(i)}_{\mathrm{img}}$.
Decoder/gating and dummy-image resampling details are provided in Appendix~\ref{app:details_decoder_gate}.

\noindent\textbf{(4) Self-distillation objective.}
Training uses anchor messages $m$ (short text strings).
The \textbf{teacher} prompt includes $m$ explicitly as text.
The \textbf{student} receives the anchor through a dummy image whose image-token span is overwritten by Eq.~\eqref{eq:vision_inject}, computed from the teacher-side rollout.
We optimize the parameters of $\mathcal{E}_i$ and $\mathcal{D}_i$ against the frozen teacher targets.

Let $h^{\mathrm{text}}$ and $\ell^{\mathrm{text}}$ denote the teacher hidden state and next-token logits at the prompt boundary, and let $h^{\mathrm{vis}}, \ell^{\mathrm{vis}}$ be the corresponding student quantities under vision injection.
We minimize
\begin{equation}
\begin{aligned}
\mathcal{L}_{\mathrm{codec}}
&= \lambda_{h}\,\bigl\|h^{\mathrm{vis}}-\mathrm{stopgrad}(h^{\mathrm{text}})\bigr\|_2^2
 + \lambda_{\mathrm{kl}}\,\tau^2\,
\mathrm{KL}\Big(
\mathrm{softmax}\!\big(\tfrac{\ell^{\mathrm{text}}}{\tau}\big)
\,\big\|\,
\mathrm{softmax}\!\big(\tfrac{\ell^{\mathrm{vis}}}{\tau}\big)
\Big) \\
&\quad + \lambda_{\mathrm{rms}}\,\Bigl(\mathrm{RMS}(\Delta_{\mathrm{inj}})-\mathrm{RMS}(\bar{X}^{(i)}_{\mathrm{img}})\Bigr)^2,
\end{aligned}
\label{eq:distill_loss}
\end{equation}
where $\tau$ is the distillation temperature and $\Delta_{\mathrm{inj}}$ denotes the gated perturbation before resampling.
The first two terms match the boundary representation and next-token distribution. The RMS term regularizes the magnitude of the injected residual.
\subsection{Aligning Codecs Across Agents via an Affine Map in Universal Space}
\label{sec:method:alignment}

Each independently trained codec defines model-specific coordinates for its $D$-dimensional tokens. We connect $N$ codecs through a hub-and-spoke design, with affine maps to and from a \emph{reference} universal space $\mathcal{U}$.
The closed-form ridge-fitting procedure is provided in Appendix~\ref{app:details_alignment}.

\noindent\textbf{Reference-space maps.}
Fix a reference agent $r$. For each agent $i$ we learn affine maps $A^{\mathrm{out}}_i(U) = U W^{\mathrm{out}}_i + \mathbf{1}(b^{\mathrm{out}}_i)^\top$ (sending: $U_i \to U^{\mathrm{ref}}$) and $A^{\mathrm{in}}_i(U) = U W^{\mathrm{in}}_i + \mathbf{1}(b^{\mathrm{in}}_i)^\top$ (receiving: $U^{\mathrm{ref}} \to U_i$), with $W^{\mathrm{out}}_i, W^{\mathrm{in}}_i \in \mathbb{R}^{D \times D}$ and $b^{\mathrm{out}}_i, b^{\mathrm{in}}_i \in \mathbb{R}^{D}$.
This yields $O(N)$ alignment maps: one map from each model to the hub and one back.

\noindent\textbf{Ridge regression on shared anchors.}
We fit these affine maps using a subset of the codec-training anchor texts, $\{m_j\}_{j=1}^M$, shared across models.
For each anchor $m_j$, we compute universal tokens $U_i(m_j)$ for every model $i$ using the already trained encoder.
We then solve a regularized least-squares problem in closed form (ridge regression) to map each model's tokens to the reference tokens:
\begin{equation}
\min_{W^{\mathrm{out}}_i,b^{\mathrm{out}}_i}\;
\sum_{j=1}^{M}\bigl\|U_i(m_j) W^{\mathrm{out}}_i + \mathbf{1}(b^{\mathrm{out}}_i)^\top - U_r(m_j)\bigr\|_F^2
+\lambda\|W^{\mathrm{out}}_i\|_F^2,
\label{eq:ridge_out}
\end{equation}
and analogously for $A^{\mathrm{in}}_i$.
The same fitted maps are reused across tasks and role assignments within a model pool.

\subsection{Inference: Multi-Agent Collaboration through the Vision Wormhole}
\label{sec:method:inference}

At inference time, intermediate agents emit universal tokens in the reference space, and the final agent generates the answer text.
Appendix~\ref{app:details_rtw} gives the full protocol.

\noindent\textbf{Message passing operator.}
A wormhole message from sender $s$ to receiver $i$ is computed as:
\begin{equation}
U^{\mathrm{ref}}_{s\rightarrow i}
= A^{\mathrm{out}}_s\!\Bigl(\mathcal{E}_s(H_s)\Bigr), \qquad
(\Delta_i, g_i)
= \mathcal{D}_i\!\Bigl(A^{\mathrm{in}}_i(U^{\mathrm{ref}}_{s\rightarrow i})\Bigr),
\label{eq:message_passing}
\end{equation}
followed by writing Eq.~\eqref{eq:vision_inject} into receiver $i$'s image-token span.

\noindent\textbf{Memory aggregation.}
Let $\mathcal{M}$ be the set of received messages for an agent, each stored as universal tokens in the reference space.
We aggregate memory by concatenation in token dimension:
$
U^{\mathrm{ref}}_{\mathrm{mem}} = \mathrm{Concat}\bigl(\{U^{\mathrm{ref}}_{m}\}_{m\in\mathcal{M}}\bigr)\in\mathbb{R}^{(|\mathcal{M}|\cdot K_{\mathrm{u}})\times D},
$
and decode a single vision-span perturbation from $U^{\mathrm{ref}}_{\mathrm{mem}}$.
The decoder cross-attends to $|\mathcal{M}|K_{\mathrm{u}}$ memory tokens and produces a fixed-length image-token span of $L^{(i)}_{\mathrm{img}}$ rows for the backbone.

\noindent\textbf{Role interaction loop.}
We follow the sequential role protocol of \citet{zou2025latent}. Each intermediate role-agent $i$ performs three operations:
    \emph{(1) Read:} Decode the current memory $U^{\mathrm{ref}}_{\mathrm{mem}}$ into a vision-span injection for agent $i$ and run the frozen backbone conditioned on this injection;
    \emph{(2) Think (latent):} Extract a rollout $H_i$ and encode it into a new universal message $U^{\mathrm{ref}}_i = A^{\mathrm{out}}_i(\mathcal{E}_i(H_i))$; and
    \emph{(3) Write:} Append $U^{\mathrm{ref}}_i$ to the shared memory buffer.
The final role generates the answer from its prompt and accumulated memory.

\section{Experiments}

\begin{figure}[t]
    \centering
    \includegraphics[width=\textwidth]{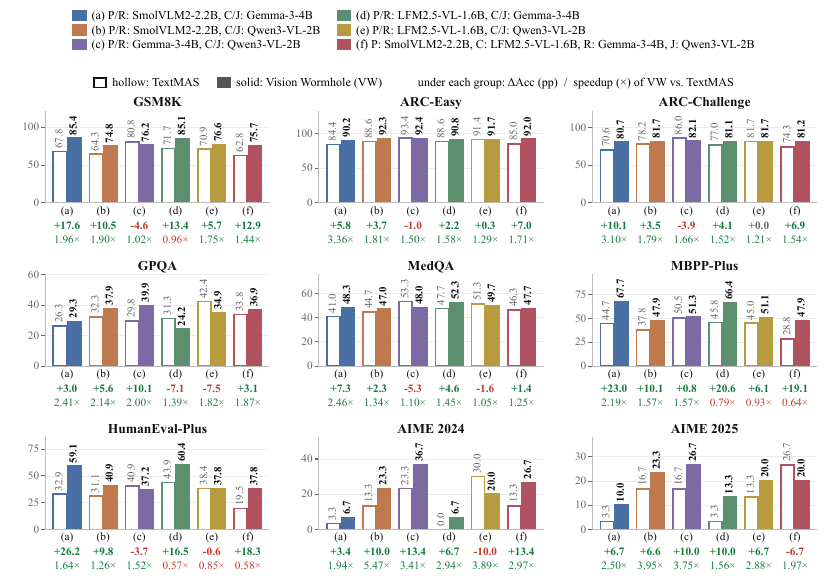}

    \caption{\textbf{Main heterogeneous MAS results: accuracy (\%).} Each panel shows one benchmark and the six configurations in Table~\ref{tab:mas_model_configs}. Hollow bars denote TextMAS and solid bars denote VW; labels below each pair give the accuracy difference in percentage points. Panels use independent $y$-axis ranges. P/C/R/J denote Planner/Critic/Refiner/Judger.}
    \label{fig:main_results}

\end{figure}

\begin{figure}[t]
    \centering
    \includegraphics[width=\textwidth]{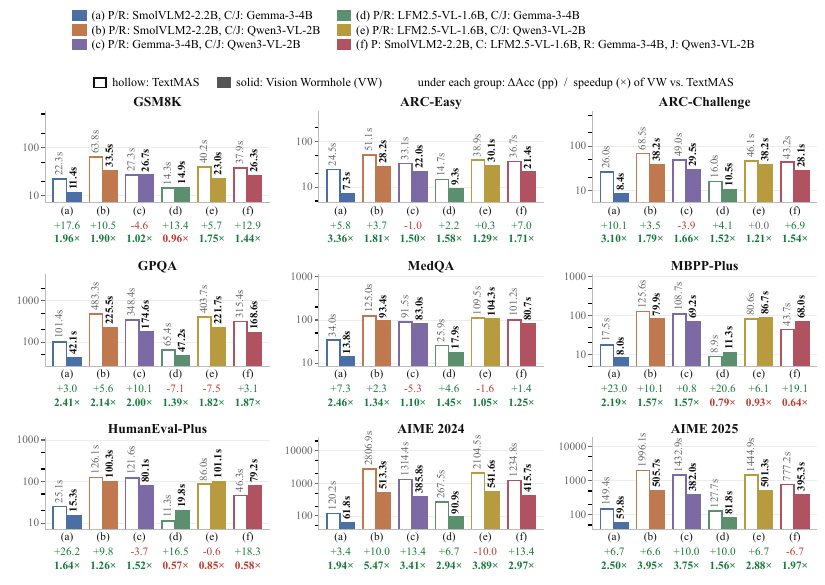}

    \caption{\textbf{Main heterogeneous MAS results: batch-normalized runtime (s/query, log scale).} Same layout, colors, and configurations as Figure~\ref{fig:main_results}; timing protocol in Appendix~\ref{app:runtime_budgets}. The number under each group is the speedup ($\times$) of VW relative to TextMAS.}
    \label{fig:main_results_time}

\end{figure}

\subsection{Experimental Settings}
\label{sec:exp_settings}

\noindent\textbf{Tasks and Datasets.}
We evaluate nine benchmarks using the evaluation sets of LatentMAS \citep{zou2025latent}: (i) \textit{Math \& Science Reasoning}, including GSM8K \citep{gsm8k} (1{,}319), AIME 2024 \citep{aime24} (30), AIME 2025 \citep{aime25} (30), GPQA Diamond \citep{gpqa} (198), and MedQA \citep{medqa} (300); (ii) \textit{Commonsense Reasoning}, including ARC-Easy (2{,}376) and ARC-Challenge (1{,}172) \citep{arc-easy,arc-challenge}; and (iii) \textit{Code Generation}, including MBPP-Plus (378) and HumanEval-Plus (164) \citep{codeplus}. Parentheses give the number of evaluated examples per configuration. We report accuracy for multiple-choice and short-answer tasks, and pass@1 for code generation.

\noindent\textbf{Models.}
We use four VLM backbones: Qwen/Qwen3-VL-2B-Thinking \citep{bai2025qwen3vltechnicalreport}, google/gemma-3-4b-it \citep{gemmateam2025gemma3technicalreport}, HuggingFaceTB/SmolVLM2-2.2B-Instruct \citep{marafioti2025smolvlmredefiningsmallefficient}, and LiquidAI/LFM2.5-VL-1.6B \citep{amini2025lfm2technicalreport}. Both TextMAS and Vision Wormhole use frozen backbones.
The main experiments use five two-backbone configurations and one four-backbone pool (model combinations and role assignments in Table~\ref{tab:mas_model_configs}, Appendix~\ref{app:impl_details}).

\noindent\textbf{Protocol, Baselines, and Implementation.}
Following \citet{zou2025latent}, all main configurations use the sequential role protocol and context accumulation in Appendix~\ref{app:details_inference}.
The primary comparison is against text-mediated MAS (TextMAS), with the same role sequence, model assignments, evaluation sets, and task-level generation caps.
We compare the visual interface with a learned text prefix (Figure~\ref{fig:visual_interface_ablation}). Further comparisons cover OCR relay (Appendix~\ref{app:ocr_baseline}) and two latent baselines on Qwen/Gemma (Appendix~\ref{app:latent_channels}).
Role prompts, hardware placement, generation budgets, and batching are in Appendix~\ref{app:runtime_budgets}.

\subsection{Results}
We report accuracy and batch-normalized end-to-end runtime for TextMAS and the Vision Wormhole (VW). Accuracy gains are in percentage points; speedup is the ratio of TextMAS runtime to VW runtime. Macro-averages weight the nine benchmarks equally. The overall accuracy gain is the arithmetic mean across all 54 configuration--task pairs, and overall speedup is the geometric mean of their runtime ratios.

\begin{figure}[t]
    \centering
    \begin{minipage}[t]{0.59\textwidth}
        \vspace{0pt}\centering
        \includegraphics[width=\linewidth]{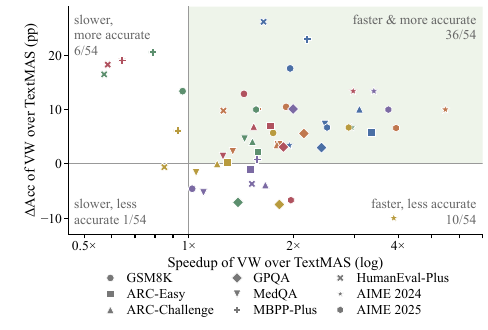}
        \vspace{-3mm}
        \caption{\raggedright\textbf{Accuracy--speed trade-off.} Points denote configuration--benchmark pairs. Colors follow Figure~\ref{fig:main_results}; markers identify benchmarks.}
        \label{fig:main_results_tradeoff}
    \end{minipage}\hfill
    \begin{minipage}[t]{0.36\textwidth}
        \vspace{0pt}\centering
        \includegraphics[width=\linewidth]{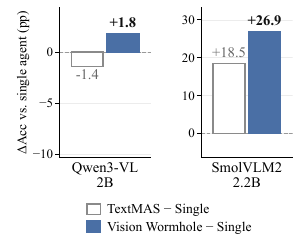}
        \vspace{-3mm}
        \caption{\raggedright\textbf{Model-group performance.} Gains over standalone (Section~\ref{sec:single_vs_mas}). Hollow: TextMAS; solid: VW. Independent $y$-axes.}
        \label{fig:single_vs_mas_summary}
    \end{minipage}
    \vspace{-4mm}
\end{figure}

\noindent\textbf{Main Heterogeneous MAS.}
VW improves accuracy by 6.0\,pp on average and achieves an overall speedup of 1.69$\times$ over TextMAS (Figures~\ref{fig:teaser}, \ref{fig:main_results}, and~\ref{fig:main_results_time}). Macro-average accuracy increases in five of six configurations, and runtime decreases in 47 of 54 configuration--task pairs. Accuracy and runtime improve together in 36 pairs (Figure~\ref{fig:main_results_tradeoff}).
Code generation has the largest benchmark-level mean accuracy gains: 13.3\,pp on MBPP-Plus and 11.1\,pp on HumanEval-Plus, averaging over the six configurations.

\noindent\textbf{Weakly Supervised Setting.}
Table~\ref{tab:results_weakly_supervised} reports results for the weakly supervised setting defined in Appendix~\ref{app:codec_training_setup}. VW retains its runtime advantage across all 18 configuration--task pairs. The SmolVLM2/Qwen configuration also improves accuracy on every benchmark.

\noindent\textbf{Performance across Model Groups.}
\label{sec:single_vs_mas}
Figure~\ref{fig:single_vs_mas_summary} compares Qwen3-VL-2B and SmolVLM2-2.2B standalone accuracy with their \emph{model-group} MAS accuracy. For each model, we average the scores of all configurations containing that model within each benchmark, then average across benchmarks.

\begin{table}[t]
    \centering
    \caption{\textbf{Weakly supervised results.} We report accuracy (\%), batch-normalized runtime (s/query), and differences relative to TextMAS ($\Delta$Acc in pp; runtime ratio in $\times$).}
    \label{tab:results_weakly_supervised}
    \scriptsize
    \setlength{\tabcolsep}{1.8pt}
    \renewcommand{\arraystretch}{1.0}
    \resizebox{\textwidth}{!}{
    \begin{tabular}{l cccccc cccccc}
        \toprule
         & \multicolumn{6}{c}{P/R: Gemma-3-4B, C/J: Qwen3-VL-2B} & \multicolumn{6}{c}{P/R: SmolVLM2-2.2B, C/J: Qwen3-VL-2B} \\
        \cmidrule(lr){2-7} \cmidrule(lr){8-13}
        \textbf{Dataset} & \textbf{Text Acc} & \textbf{Text Time} & \textbf{VW Acc} & \textbf{VW Time} & \textbf{$\Delta$Acc} & \textbf{Speedup} & \textbf{Text Acc} & \textbf{Text Time} & \textbf{VW Acc} & \textbf{VW Time} & \textbf{$\Delta$Acc} & \textbf{Speedup} \\
        \midrule
        GSM8K & 80.8\% & 27.3s & 77.6\% & 22.9s & \textcolor{red!70!black}{-3.2pp} & \textcolor{green!60!black}{$1.19\times$} & 64.3\% & 63.8s & 77.0\% & 25.6s & \textcolor{green!60!black}{+12.7pp} & \textcolor{green!60!black}{$2.49\times$} \\
        ARC-Easy & 93.4\% & 33.1s & 90.3\% & 23.1s & \textcolor{red!70!black}{-3.1pp} & \textcolor{green!60!black}{$1.43\times$} & 88.6\% & 51.1s & 91.8\% & 23.6s & \textcolor{green!60!black}{+3.2pp} & \textcolor{green!60!black}{$2.17\times$} \\
        ARC-Challenge & 86.0\% & 49.0s & 80.5\% & 30.3s & \textcolor{red!70!black}{-5.5pp} & \textcolor{green!60!black}{$1.62\times$} & 78.2\% & 68.5s & 81.3\% & 30.4s & \textcolor{green!60!black}{+3.1pp} & \textcolor{green!60!black}{$2.25\times$} \\
        GPQA & 29.8\% & 348.4s & 34.9\% & 172.9s & \textcolor{green!60!black}{+5.1pp} & \textcolor{green!60!black}{$2.02\times$} & 32.3\% & 483.3s & 33.3\% & 172.5s & \textcolor{green!60!black}{+1.0pp} & \textcolor{green!60!black}{$2.80\times$} \\
        MedQA & 53.3\% & 91.5s & 45.0\% & 82.0s & \textcolor{red!70!black}{-8.3pp} & \textcolor{green!60!black}{$1.12\times$} & 44.7\% & 125.0s & 49.0\% & 75.8s & \textcolor{green!60!black}{+4.3pp} & \textcolor{green!60!black}{$1.65\times$} \\
        MBPP-Plus & 50.5\% & 108.7s & 46.6\% & 69.1s & \textcolor{red!70!black}{-3.9pp} & \textcolor{green!60!black}{$1.57\times$} & 37.8\% & 125.6s & 49.2\% & 69.4s & \textcolor{green!60!black}{+11.4pp} & \textcolor{green!60!black}{$1.81\times$} \\
        HumanEval-Plus & 40.9\% & 121.6s & 40.2\% & 80.7s & \textcolor{red!70!black}{-0.7pp} & \textcolor{green!60!black}{$1.51\times$} & 31.1\% & 126.1s & 42.7\% & 80.6s & \textcolor{green!60!black}{+11.6pp} & \textcolor{green!60!black}{$1.56\times$} \\
        AIME 2024 & 23.3\% & 1314.4s & 26.7\% & 404.8s & \textcolor{green!60!black}{+3.4pp} & \textcolor{green!60!black}{$3.25\times$} & 13.3\% & 2806.9s & 36.7\% & 389.8s & \textcolor{green!60!black}{+23.4pp} & \textcolor{green!60!black}{$7.20\times$} \\
        AIME 2025 & 16.7\% & 1432.9s & 23.3\% & 370.5s & \textcolor{green!60!black}{+6.6pp} & \textcolor{green!60!black}{$3.87\times$} & 16.7\% & 1996.1s & 23.3\% & 415.8s & \textcolor{green!60!black}{+6.6pp} & \textcolor{green!60!black}{$4.80\times$} \\
        \bottomrule
    \end{tabular}
    }

\end{table}

VW improves on TextMAS in both model groups.

\begin{wrapfigure}[11]{r}{0.60\textwidth}
    \centering
    \includegraphics[width=\linewidth]{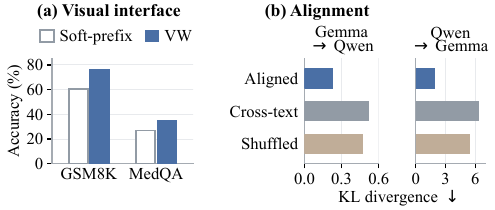}
    \caption{\textbf{Interface and alignment ablations.} (a) Qwen accuracy. (b) Held-out reconstruction KL.}
    \label{fig:visual_interface_ablation}
\end{wrapfigure}
\noindent\textbf{Component Ablations.}
Visual reception outperforms a learned soft prefix on GSM8K and MedQA with matched Qwen codec training and inputs (Figure~\ref{fig:visual_interface_ablation}a). For Qwen/Gemma ablation codecs, aligned messages give lower reconstruction error on held-out texts than cross-text messages or shuffled-anchor maps in both directions (Figure~\ref{fig:visual_interface_ablation}b). Appendices~\ref{app:native_interfaces} and~\ref{app:heldout_reconstruction} give the setups and full results.

\section{Conclusion}

Vision Wormhole turns the visual input interface of a VLM into a channel for receiving another model's internal states. Per-model codecs encode latent rollouts into universal messages and decode incoming messages into visual context. Self-distillation trains each codec with its backbone frozen; affine reference maps connect the codecs with $O(N)$ components for $N$ models. Across four VLM families and nine benchmarks, VW improves accuracy by 6.0 percentage points on average and achieves a 1.69$\times$ overall speedup over TextMAS. These results establish visual conditioning as a practical interface for continuous communication among frozen heterogeneous models.

\subsection*{AI Use Statement}

We used large language models as general-purpose assistants for literature search, manuscript drafting and editing, experimental design, code implementation, result analysis, and figure preparation. Language models also supported the evaluations in Appendices~\ref{app:latentmas_hybrid_failure} and~\ref{app:sender_quality}. The authors reviewed the AI-assisted work and take responsibility for the final manuscript and results.

\subsection*{Ethics Statement}

The experiments study communication among publicly released language and vision-language models using public reasoning and code-generation benchmarks. The system inherits the capabilities and behavioral limitations of those checkpoints. Evaluation is confined to the benchmark tasks described in Section~\ref{sec:exp_settings}.

\subsection*{Reproducibility Statement}

The codec architecture, training objective, and alignment procedure are specified in Section~\ref{sec:method}, with training hyperparameters, anchor-corpus construction, and optimizer settings in Appendix~\ref{app:codec_training_setup} and architectural details in Appendix~\ref{app:details_codec}. Hardware placement, generation budgets, decoding, prompt sources, and batching are given in Appendix~\ref{app:runtime_budgets}. Model configurations and role assignments are listed in Table~\ref{tab:mas_model_configs}. We use publicly released benchmarks and model checkpoints, with the LatentMAS evaluation sets listed in Section~\ref{sec:exp_settings}.

\clearpage
\appendix

\section{Extended Related Work}
\label{app:related_work}

\subsection{Multi-Agent Workflows, Memory, and Tools}
Surveys of LLM-based multi-agent systems catalogue role assignments, coordination patterns, and evaluation practices \citep{guo2024large,tran2025multi,yan2025beyond,yang2024llm,acharya2025agentic,wang2024survey,yao2025survey,li2025embodied,feng2025multi,zhang2024large,zhao2025llm}. Communication surveys connect these systems to learned communication in multi-agent reinforcement learning and emergent language through the choice of participants, message content, timing, and purpose \citep{chen2026wsmultiagentcommunicationtalks}. Agent frameworks implement natural-language interaction together with routing, delegation, and tool calls \citep{wu2024autogen,hong2023metagpt,li2023camel,fourney2024magentic,hou2025halo,hu2025owl,tao2024magis,park2023generative,wu2025talk,wang2025talk}. Workflow research studies multi-step collaboration, self-improvement, and allocation of inference computation \citep{zhang2024chain,zhao2025connecting,zhao2025sirius,chen2024optima,chen2025optima,zhou2025reso,li2025flow,zhuge2024language}. Graph-based orchestration also coordinates specialized agents through a centralized workspace \citep{hao2026braininspiredgraphmultiagentsystems}. Analyses of agent systems identify information sharing, coordination, and specialization as contributors to task outcomes \citep{pezeshkpour2024reasoning,cemri2025multi,pappu2026multiagentteamsholdexperts,zhang2025stopovervaluingmultiagentdebate}.

Memory and planning modules maintain information across interactions and support long-horizon execution \citep{zhang2025survey,du2025rethinking,packer2023memgpt,zhong2024memorybank,wang2024memoryllm,hu2025hiagent,pan2025secom}. Tool-use methods add retrieval and action selection to these workflows \citep{ruan2023tptu,zhou2024languageagenttreesearch,liu2023llmpempoweringlargelanguage,birr2024autogptpaffordancebasedtaskplanning,anantha2023protipprogressivetoolretrieval,zhuang2023toolchainefficientactionspace}. Our experiments use a sequential four-role workflow to evaluate a continuous communication interface among heterogeneous VLMs.

Heterogeneous model selection connects these workflows to differences in agent capabilities. Function--domain evaluations compare models on question answering, revision, aggregation, planning, and evaluation, then use those measurements to assign models within existing agent topologies \citep{ye2025xmasbuildingmultiagentsystems}. Layered response aggregation studies separate proposer and aggregator roles and finds benefits from diverse model proposals in its generation benchmarks \citep{wang2024mixtureofagentsenhanceslargelanguage}. These results motivate communication interfaces that can serve several model families within one workflow.

\subsection{Continuous Messages and Hidden-State Interfaces}
Latent communication exposes intermediate model computations as messages. A recent survey organizes these interfaces by the representation transferred, the sender--receiver alignment, and the receiver's fusion operation \citep{liu2026tokensunifiedframeworklatent}. These choices connect the message format to concrete implementation requirements: its token axis, feature dimension, layer correspondence, and insertion point. Probability-weighted vocabulary embeddings provide a continuous representation of the sender's next-token distribution \citep{pham2024cipher}. Activation communication inserts a sender's intermediate vector into the receiver's residual stream, with arithmetic fusion and an optional model-pair alignment matrix \citep{ramesh2025activations}. State-delta communication augments generated text with token-wise changes in internal states \citep{tang2025augmentingmultiagentcommunicationstate}. Together, these approaches establish several ways to expose information from the sender's forward computation. Latent communication has also been examined in dedicated evaluation studies \citep{wenzel2026latentcommunicationlanguagemodel,lu2026latentrepresent,lu2026alignmentrole,zhang2026latentchannelsactuallycommunicate,cheng2026doeslatentcommunicationpay}.

Input-level interfaces place translated messages alongside the receiver's prompt embeddings. Learned adapters support the transfer and compression of last-layer state sequences \citep{du-etal-2026-enabling}, and vector translators align representations between independently trained language models \citep{yang2025directsemanticcommunicationlarge}. Hybrid lexical--latent interfaces combine receiver-native embeddings of the sender's context tokens with gated cross-attention over its hidden states \citep{yang2026xbridgeentitygroundedlatentbridge}. These designs give concrete roles to contextual representations, lexical identity, and input-space calibration.

\subsection{KV-Cache Communication and Heterogeneous Translation}
KV-cache interfaces transfer attention state across model calls. Selective sharing chooses sender layers using attention importance and a layer prior \citep{shi2026kvcommenablingefficientllm}. Latent collaboration accumulates intermediate computation in a shared working-memory cache and passes this state between role-specific agents \citep{zou2025latent}. Reusable latent primitives compose recurring review, voting, and planning operations with internal KV-cache communication \citep{jin2026agentprimitivesreusablelatent}. These methods link the granularity of the transmitted state to the organization of the collaborative workflow.

Heterogeneous cache translation learns how sender tensors condition a receiver with its own representation geometry. Cache-fusion networks project source and target states into a receiver-side update and learn gates for the participating layers \citep{fu2025cache}. Joint key--value translation and pooling produce compact updates for both shared-context and different-context exchanges \citep{rossi2026latentcacheflowmodeltomodel}. Dense cache alignment uses reconstruction followed by generation-oriented training to transfer source context across models with different depths and KV layouts \citep{chen2026iseeknowi}. Its evaluation includes both receivers that observe the original context and receivers whose source information arrives through the cache. This line of work develops explicit transformations at the receiver's attention-state interface.

\subsection{Shared Spaces, Model Composition, and Communication Training}
Learned shared spaces make model composition a reusable operation. Cross-attention modules connect frozen language models so that one model can draw on representations produced by another \citep{bansal2024llmaugmentedllmsexpanding}. Shared-space expert collaboration combines query-dependent routing with interaction layers that project selected experts' hidden states into a common space \citep{fein2025mixture}. Latent-variable approaches use sparsity-regularized reconstruction to infer shared and private components of agent states and route them through prefix adaptation \citep{zheng2025thought}. These works establish shared representations as a design for composing model capabilities.

Communication objectives also vary with the intended interaction. Parameter-efficient supervised training over multi-agent latent trajectories jointly optimizes how information is produced and consumed across successive roles \citep{yu2026learningcommunicateendtoendoptimization}. Reconstruction-guided test-time optimization uses a question--latent--question cycle to update latent states before answer generation \citep{yuan2026closinglooplatentreasoning}. Hybrid diffusion--autoregressive systems train a projector from a diffusion planner's representations to an autoregressive executor's input embeddings \citep{berrayana2026latentdarmbridgingdiscretediffusion}. The Vision Wormhole trains each visual codec from its backbone's prompt-boundary state and next-token distribution, then fits reference maps on shared anchor texts. Its shared-space messages connect separately trained codecs at the visual input interface.

\subsection{Message Compression and Cache Execution}
Communication compression controls the amount of state stored and transported. Adaptive quantization allocates precision across KV layers and combines compressed caches with explicit fact extraction and model calibration \citep{kriuk2025qkvcommefficientmultiagentcommunication}. Cache-relay compression retains selected KV entries and folds a low-rank residual from discarded states into the retained representation \citep{li2026latentleadsbetterrelay}. Post-hoc sparse coding has also been applied to frozen Vision Wormhole messages, measuring payload size, reconstruction, downstream accuracy, and feature reuse \citep{wu2026posthocsparsecodinglatent}. These approaches operate on different message representations and compression stages; their relevant budgets include scalar precision, token or layer count, and the storage required for the coding mechanism.

Execution systems reuse attention state across agents and conversation phases. Online cross-context reuse estimates cache offsets from an adaptive anchor pool \citep{ye2025kvcomm}; decoding-cache relay selectively recomputes positions affected by prefix changes \citep{geng2026relaycachingacceleratingllmcollaboration}. Multi-LoRA caching decomposes attention state into shared backbone and low-rank adapter components \citep{jeon2026lragentefficientkvcache}. Collective cache sharing exploits repeated output blocks across synchronized agent rounds and stores differences relative to a shared copy \citep{bian2026tokendancescalingmultiagentllm}. Persistent quantized caches support agent-context restoration on memory-constrained devices \citep{shkolnikov2026agentmemorypromptpersistent}, while distributed workflow operators manage cache materialization, transfer, composition, and eviction \citep{sarker2026aaflowstatefuloperatorabstraction}. Network-aware studies jointly schedule cache delivery and token prefill, or select between token and KV transport under computation and bandwidth constraints \citep{lee2026lowlatencyedgellmhandover,dai2026tokenkvcachecommunicationmediaselection}. This systems work makes transport and receiver execution explicit parts of the communication cost.

\subsection{Multimodal and Domain-Specific Collaboration}
Latent communication is also being developed for agents with distinct observations and execution budgets. Visual agent systems maintain separate perception and reasoning memories with entropy-dependent access \citep{yu2026duallatentmemoryvisual}. Long-video collaboration assigns video segments to local agents, compresses their observations into shared-space tokens, and trains a coordinator through a staged curriculum \citep{chen2026scalingvideounderstandingcompact}. Continuous slow--fast coupling projects a deliberative model's residuals into a fast model's input embeddings for real-time action selection \citep{li2026latentbridgecontinuousslowfast}. Collaborative driving uses iterative latent deliberation, saliency-based state selection, and structured distillation to support decisions from partial observations \citep{chen2026lacoadaptivelatentcommunication}. In cross-hospital diagnosis, local agents distill clinical evidence into compact KV blocks, with separate alignment procedures for shared and heterogeneous backbones \citep{wang2026medlatentdxlatentmultiagentcommunication}. These applications connect message design to the division of observations, roles, and computation among agents.

\subsection{Security of Latent Channels}
Security research studies the handling of communicated state. Latent interventions test how attack-associated information propagates through hidden states and KV handoffs \citep{wang2026sightmindunveilinglatent}. Cache-integrity studies bind transmitted tensors to sender, session, and message metadata through authenticated manifests \citep{brito2026latentagentsliekvcache}. Adversarial representation learning trains cache transformations to preserve task information while reducing reconstruction of sensitive source inputs \citep{asif2026lcguardlatentcommunicationguard}. This work develops auditing, integrity, and disclosure controls for latent communication interfaces.

\subsection{Continuous Reasoning and Internalized Collaboration}
Continuous reasoning methods study how a model can perform intermediate computation through recurrent hidden states or continuous token representations \citep{hao2024training,zhang2025soft,zhu2025reasoning,liu2024deliberation,coda2025exploring,qu2025survey,sui2025stop,wang2025harnessing,zhang2025surveyrein}. Internalized collaboration distills multi-agent debate into a single model and studies the resulting agent-specific activation directions \citep{yi2026latentagentsposttrainingprocedure}. Related steering approaches learn a codebook of latent roles and compose them with a query-dependent router \citep{zeng2026modelmindsunlockingmultiagent}. The Vision Wormhole uses a cached latent rollout as the input to a communication encoder. Its receiver conditions on the decoded message while performing its own role-specific computation. Appendix~\ref{app:latent_channels} specifies the implementation sources, tokenizer handling, and role sequences used to evaluate earlier communication channels.

\subsection{Cross-Model Representation Alignment}
Tokenizer transfer, vocabulary alignment, and cross-tokenizer distillation study compatibility between models with different token inventories \citep{minixhofer2024zero,feher2025retrofitting,minixhofer2025universal,li2025tokalign,remy2023tik,remy2024trans,tai2020exbert,pfeiffer2021unks,rust2021good,vernikos2021subword,mundra2024empirical,yamaguchi2024empirical,yamaguchi2025how,moroni2025optimizing,downey2023embedding,goddard2025training,sharthak2025achieving,liu2026trojanvocabularystealthysabotage}. At the representation level, linear alignment, model stitching, and permutation-aware merging investigate correspondences between independently trained networks \citep{bansal2021revisiting,lenc2015equivariance,ainsworth2022git,wortsman2022model,park2023linear,zou2023representation,zhou2019improving}. Research on relative representations and latent communication studies how semantic anchors support reuse across neural spaces and modalities \citep{moschella2024latentcommunicationartificialneural}. We use affine regression on corresponding anchor representations to connect each codec to a chosen reference model. The encoder, decoder, and reference maps form a reusable interface for each backbone within the model pool.

\subsection{Continuous Prompts and Multimodal Conditioning}
Prompt Tuning and Prefix-Tuning demonstrate learned continuous conditioning of frozen language models \citep{lester2021prompttuning,li2021prefixtuning}. Vision-language architectures condition language generation on image features processed by learned projectors or resamplers \citep{alayrac2022flamingo,li2023blip2}. Contrastive image-text training studies shared representations across modalities \citep{radford2021clip,jia2021align,zhai2023siglip}, and recent work examines alignment and reasoning within multimodal hidden spaces \citep{fein2025bridging,li2025latent,wang2025monet}.

The Vision Wormhole generates a continuous prompt in a receiver's image-token span from a sender's latent rollout. Its per-model self-distillation objective matches the text-conditioned backbone at the prompt boundary, and reference maps compose the resulting codecs for heterogeneous communication. This combination links continuous conditioning, cross-model alignment, and multi-agent inference in one modular architecture.

\section{Implementation Details}
\label{app:impl_details}

\subsection{Main Heterogeneous MAS Model Configurations}

Table~\ref{tab:mas_model_configs} lists the heterogeneous MAS model configurations used in the main experiments.

\begin{table}[!h]
    \centering
\caption{\textbf{Main heterogeneous MAS model configurations.}
    Two-backbone configurations alternate backbones across roles.}
    \label{tab:mas_model_configs}
    \resizebox{0.98\textwidth}{!}{
    \begin{tabular}{lll}
        \toprule
        \textbf{Backbone Setup} & \textbf{Backbones} & \textbf{Role assignment (P/C/R/J)} \\
        \midrule
        \multicolumn{3}{l}{\textit{Two-backbone configurations}} \\
        2 backbones & Gemma-3-4B + SmolVLM2-2.2B & (SmolVLM2, Gemma-3-4B, SmolVLM2, Gemma-3-4B) \\
        2 backbones & Qwen3-VL-2B + SmolVLM2-2.2B & (SmolVLM2, Qwen3-VL-2B, SmolVLM2, Qwen3-VL-2B) \\
        2 backbones & Qwen3-VL-2B + Gemma-3-4B & (Gemma-3-4B, Qwen3-VL-2B, Gemma-3-4B, Qwen3-VL-2B) \\
        2 backbones & LFM2.5-VL-1.6B + Gemma-3-4B & (LFM2.5-VL-1.6B, Gemma-3-4B, LFM2.5-VL-1.6B, Gemma-3-4B) \\
        2 backbones & LFM2.5-VL-1.6B + Qwen3-VL-2B & (LFM2.5-VL-1.6B, Qwen3-VL-2B, LFM2.5-VL-1.6B, Qwen3-VL-2B) \\
        \midrule
        \multicolumn{3}{l}{\textit{Four-backbone pool (1.6B--4B)}} \\
        4 backbones & \begin{tabular}[t]{@{}l@{}}SmolVLM2-2.2B + LFM2.5-VL-1.6B +\\ Gemma-3-4B + Qwen3-VL-2B\end{tabular} & (SmolVLM2, LFM2.5-VL-1.6B, Gemma-3-4B, Qwen3-VL-2B) \\
        \bottomrule
    \end{tabular}
    }

\end{table}

\subsection{Codec Training Setup (Shared Across Runs)}
\label{app:codec_training_setup}

\paragraph{Objective.}
We train a latent-to-vision injection codec for each backbone model.
Multi-model inference combines these per-model codecs with fitted reference-space maps.

\paragraph{Anchor corpora.}
We consider two anchor settings:
(i) the \textbf{default} setting uses 3,000 texts drawn equally from \texttt{cos\_e}, \texttt{OpenCodeReasoning}, and \texttt{PRM800K};
and (ii) the \textbf{weakly supervised} setting uses 90 texts, with equal contributions from the same sources.

\paragraph{Anchor construction.}
For each data source, we cap the number of sampled examples, concatenate sources, and shuffle deterministically.

\paragraph{Step-based random sampling.}
At each optimization step, we sample a mini-batch uniformly with replacement from the anchor pool.

\paragraph{Codec architecture hyperparameters.}
We use universal dimension $D=512$ and $K=1024$ learned message queries. A global query and a style query give $K_{\mathrm{u}}=1026$ total tokens per message. The decoder uses $K_{\mathrm{img}}=256$ image queries, with its output resampled to the receiver's image-span length.
The codec uses 6 transformer layers with 8 attention heads and dropout 0.10.
Codec training uses $T_{\mathrm{train}}=1024$ latent steps.

\paragraph{Optimization and losses.}
We optimize with AdamW at a learning rate of $2\times 10^{-4}$ for 400 steps with batch size 2.
The training objective is a weighted sum of three terms:
\begin{itemize}[leftmargin=*, nosep]
    \item Hidden-state MSE (weight 1.0).
    \item Logit alignment via KL divergence (weight 0.25; temperature 1.0).
    \item Injection-statistics regularizer (weight 0.1).
\end{itemize}
We apply gradient clipping with max-norm 1.0 and use standard numerical stabilization, including clipping for latent/logit/injection values and non-finite guards.

\paragraph{Rollout computation.}
Each training example produces one latent rollout from its teacher prompt. The frozen backbone feeds normalized hidden states back as continuous pseudo-tokens.

\paragraph{What changes across variants.}
Across codec variants, we vary the backbone model, the anchor corpus size, and the merge pairing used at inference.
We keep the core codec architecture, optimizer/loss weights, step count, batch size, and the step-based sampling strategy fixed.

\paragraph{Merged codecs.}
For multi-model inference, we assemble the trained per-model codecs and fit universal-space alignment by ridge regression on texts from the codec-training pool. The default setting randomly selects 300 alignment texts; the weakly supervised setting reuses the entire training pool. The Qwen3-VL/Gemma-3 configuration uses Qwen3-VL as reference and ridge coefficient $\lambda=10^{-3}$.

\subsection{Experiment Runtime and Generation Budgets}
\label{app:runtime_budgets}

\paragraph{Decoding and evaluation.}
The main experiments use sampling with temperature 0.6, top-$p$ 0.95, and seed 42. Both methods use the task-specific generation caps in Table~\ref{tab:gen_budgets}.

\paragraph{Role instructions.}
We use the text- and latent-context role prompts of \citet{zou2025latent}, with backbone-specific system messages and benchmark-specific answer formats. VW carries the latent context through the visual interface.

\paragraph{Hardware and model placement.}
All experiments are conducted on NVIDIA A6000 GPUs.
For two-backbone runs in the main setting, we colocate both backbones on a single GPU.
For the four-backbone pool, we use two GPUs and place half of the agents on each GPU.

\paragraph{Per-dataset token budgets.}
Table~\ref{tab:gen_budgets} lists the shared generation caps and default batch sizes.

\begin{table}[!h]
    \centering
    \caption{\textbf{Per-dataset generation budgets and default batch sizes.}
    We adopt the same maximum generation budgets as \citet{zou2025latent}.}
    \label{tab:gen_budgets}
    \small
    \begin{tabular}{@{}lcc@{}}
        \toprule
        \textbf{Dataset} & \makecell{\textbf{Max. new}\\\textbf{tokens}} & \makecell{\textbf{Default}\\\textbf{batch size}} \\
        \midrule
        GSM8K & 2048 & 12 \\
        ARC-Easy & 2048 & 12 \\
        ARC-Challenge & 2048 & 12 \\
        MedQA & 4096 & 8 \\
        MBPP-Plus & 4096 & 8 \\
        HumanEval-Plus & 4096 & 8 \\
        GPQA & 8192 & 4 \\
        AIME 2024 & 20000 & 4 \\
        AIME 2025 & 20000 & 4 \\
        \bottomrule
    \end{tabular}

\end{table}

\paragraph{Dynamic batching and time reporting.}
All methods in the main experiments use the default batch sizes in Table~\ref{tab:gen_budgets} and the same out-of-memory (OOM) backoff schedule, \{12, 8, 4, 2, 1\}. Starting from the dataset-specific default, we reduce the batch size along this schedule until the run succeeds. We report batch-normalized runtime (seconds/query) as the total wall-clock time across evaluation batches divided by the number of evaluated examples. Timing includes intermediate-role computation, message processing, and final-answer generation.

\section{Additional Details: Codec Training, Alignment, and Inference}
\label{app:details_codec}

This appendix specifies the codec architecture, alignment fit, and inference procedure.

\subsection{Training a Codec for a Specific VLM}
\label{app:details_codec_train}

\paragraph{Pseudo-token normalization.}
Latent rollouts feed back continuous pseudo-token embeddings derived from hidden states.
A per-model normalization operator sets their magnitude relative to the token-embedding scale:
\begin{equation}
\mathrm{NormMatch}_i(h) \;=\; \alpha_i \cdot \frac{h}{\|h\|_2 + \epsilon},
\label{eq:normmatch}
\end{equation}
where $\alpha_i$ is the typical token-embedding norm for model $i$, e.g.,
$\alpha_i = \mathbb{E}_{w\sim\mathcal{V}_i}\|E_i(w)\|_2$, and $\epsilon$ is a small constant.
The normalized pseudo-token has norm approximately $\alpha_i$.

\paragraph{Latent rollout with cached context.}
\label{app:details_rollout_cached}
Let the prompt produce cached attention keys/values (equivalently, a fixed conditioning context).
A rollout step appends a single pseudo-token embedding $x_t$ and computes the next hidden state at the new position.
Repeating for $T$ steps yields $H_i\in\mathbb{R}^{T\times d_i}$.
The rollout length $T$ is a fixed hyperparameter that bounds message extraction cost.

\paragraph{Perceiver resampler encoder.}
\label{app:details_resampler}
The encoder $\mathcal{E}_i$ maps $H_i$ to $K_{\mathrm{u}}$ tokens of dimension $D$.
We first project the rollout into the universal dimension:
\begin{equation}
Z = H_i P_i \in \mathbb{R}^{T \times D},
\end{equation}
for a learned matrix $P_i\in\mathbb{R}^{d_i\times D}$.
We then maintain a fixed set of learned queries
$Q^{(0)}\in\mathbb{R}^{K_{\mathrm{u}}\times D}$ and update them through $L$ cross-attention blocks:
\begin{align}
\tilde{Q}^{(\ell+1)}
&= Q^{(\ell)} + \mathrm{MHA}\!\left(\mathrm{LN}(Q^{(\ell)}), \mathrm{LN}(Z), \mathrm{LN}(Z)\right), \\
Q^{(\ell+1)}
&= \tilde{Q}^{(\ell+1)} + \mathrm{FFN}\!\left(\mathrm{LN}(\tilde{Q}^{(\ell+1)})\right),
\end{align}
for $\ell = 0, \dots, L-1$.
This is the Perceiver resampler pattern \citep{alayrac2022flamingo}: a constant number of queries attends to a variable-length latent sequence.

\paragraph{Global and style tokens.}
\label{app:details_global_style}
In addition to the $K$ message tokens, we include a global token (for pooling) and a style token (for scale/statistics cues).
The style statistic vector is
\begin{equation}
s(H_i)
=
\left[
\mathrm{mean}(H_i),\,
\mathrm{std}(H_i),\,
\frac{1}{T}\sum_{t=1}^{T}\|H_{i,t}\|_2
\right]
\in \mathbb{R}^{3},
\label{eq:style_stats}
\end{equation}
which is mapped by a small MLP into $\mathbb{R}^{D}$ and added to the style token.
The style token thus includes the rollout's mean, standard deviation, and average vector norm.

\paragraph{Universal-to-vision decoder.}
\label{app:details_decoder_gate}
The decoder $\mathcal{D}_i$ mirrors the resampler pattern.
A learned set of $K_{\mathrm{img}}$ image queries attends to the (possibly concatenated) universal tokens to produce $K_{\mathrm{img}}$ vectors, which are linearly projected into $\mathbb{R}^{d_i}$ to form $\Delta_i$.
A gate $g_i\in(0,1)$ is predicted from a pooled representation of the universal tokens.
The gate sets a learned, example-dependent scale for the residual.

\paragraph{Dummy-image baseline and length resampling.}
\label{app:details_dummy_resample}
Different VLMs use different image-token lengths $L^{(i)}_{\mathrm{img}}$.
We compute once per model a baseline image-span embedding $\bar{X}^{(i)}_{\mathrm{img}}$ using a fixed dummy image.
At inference, we resample the decoded $\Delta_i \in \mathbb{R}^{K_{\mathrm{img}}\times d_i}$ to the required span length
$\mathrm{Resample}(\Delta_i;L^{(i)}_{\mathrm{img}})\in\mathbb{R}^{L^{(i)}_{\mathrm{img}}\times d_i}$
using linear interpolation along the token index.
The injected span is written as in Eq.~\eqref{eq:vision_inject}.

\paragraph{Distillation signals and where they are taken.}
\label{app:details_distill_boundary}
The teacher processes the anchor text $m$ and provides two targets at the prompt boundary: a hidden-state vector and a distribution over the next token. The teacher-side rollout, initialized from this boundary, supplies the encoder input. The student processes the corresponding vision-conditioned prompt, and the loss matches its boundary state and next-token distribution to these targets.

\subsection{Affine Alignment in Universal Space}
\label{app:details_alignment}

\paragraph{Reference-space construction.}
Each encoder produces tokens in its learned coordinates. We fit affine maps between each model's tokens and those of the reference model, using the same subset of codec-training texts and matching query positions.

\paragraph{Closed-form ridge regression.}
\label{app:details_ridge}
Given anchor texts $\{m_j\}_{j=1}^M$, we obtain token matrices $U_i(m_j)$ and $U_r(m_j)$ for model $i$ and reference $r$.
We flatten across anchors and token positions to form
$X_i \in \mathbb{R}^{(M K_{\mathrm{u}})\times D}$ and $Y_r\in\mathbb{R}^{(M K_{\mathrm{u}})\times D}$,
and solve
\begin{equation}
\min_{W,b}\;\;\|X_i W + \mathbf{1}b^\top - Y_r\|_F^2 + \lambda\|W\|_F^2,
\end{equation}
Let $X_c$ and $Y_c$ denote the column-centered matrices, with column means $\bar{x}$ and $\bar{y}$. The solution is
\begin{equation}
W=(X_c^\top X_c+\lambda I)^{-1}X_c^\top Y_c,
\qquad b=\bar{y}-W^\top\bar{x}.
\end{equation}
We fit the forward map $A^{\mathrm{out}}_i$ and reverse map $A^{\mathrm{in}}_i$ separately, exchanging the regression inputs and targets.

\paragraph{Component count.}
For a fixed reference, each additional model contributes one encoder, one decoder, and two affine maps. The reference model uses identity maps. Thus, the architecture contains $N$ codecs and $2(N-1)$ fitted maps.

\subsection{Inference Protocols and Role Interaction}
\label{app:details_inference}

\paragraph{Unified ``read--think--write'' abstraction.}
\label{app:details_rtw}
The sequential workflow applies the following operations to each intermediate role:
\begin{enumerate}[leftmargin=*, nosep]
    \item \textbf{Read memory:} decode aggregated $U^{\mathrm{ref}}_{\mathrm{mem}}$ into an injected image span for the current agent.
    \item \textbf{Think:} run the frozen backbone conditioned on the injection to produce either
    (a) a latent rollout $H_i$ (for intermediate roles), or
    (b) an output token sequence (for the final role).
    \item \textbf{Write memory:} encode $H_i$ into a new universal message $U^{\mathrm{ref}}_i$ and add it to memory.
\end{enumerate}

\paragraph{Accumulated context.}
\label{app:details_collab_modes}
We follow \citet{zou2025latent} for sequential context accumulation. The Planner, Critic, and Refiner execute in order, each appending one message to memory; the Judger generates the final answer after reading the accumulated messages.

\paragraph{Message processing.}
\label{app:details_mem_budget}
Reading $|\mathcal{M}|$ messages involves decoder cross-attention over $|\mathcal{M}|K_{\mathrm{u}}$ tokens, followed by a fixed-length image span of $L^{(i)}_{\mathrm{img}}$ tokens in the backbone. Sender computation includes prompt prefill, a cached latent rollout, and codec encoding.

\section{OCR-Based Image Relay Baseline}
\label{app:ocr_baseline}

The OCR relay sends natural-language messages as rendered images within the four-role workflow. Each sender generates a text message, a renderer converts it to an image, and the receiver reads the image through its standard visual input.

Table~\ref{tab:results_ocr_baseline} compares the OCR relay with the TextMAS and VW runs from Figures~\ref{fig:main_results} and~\ref{fig:main_results_time} on Gemma/Qwen and SmolVLM2/Qwen. VW achieves higher accuracy than the OCR relay on all nine benchmarks in both configurations, together with lower macro-average runtime in each configuration.

\begin{table}[t]
    \centering
    \caption{\textbf{OCR-based image-relay baseline.} OCR renders the sender's generated text as an image, which the receiver reads through its native visual input. Each cell reports accuracy (\%) and average wall-clock time (s/query).}
    \label{tab:results_ocr_baseline}
    \scriptsize
    \setlength{\tabcolsep}{4pt}
    \renewcommand{\arraystretch}{1.15}
    \begin{tabular}{l ccc ccc}
        \toprule
         & \multicolumn{3}{c}{\shortstack{P/R: Gemma-3-4B\\C/J: Qwen3-VL-2B}} & \multicolumn{3}{c}{\shortstack{P/R: SmolVLM2-2.2B\\C/J: Qwen3-VL-2B}} \\
        \cmidrule(lr){2-4} \cmidrule(lr){5-7}
        \textbf{Dataset} & \textbf{Text} & \textbf{VW} & \textbf{OCR} & \textbf{Text} & \textbf{VW} & \textbf{OCR} \\
        \midrule
        GSM8K & 80.8\% / 27.3s & 76.2\% / 26.7s & 72.8\% / 112.9s & 64.3\% / 63.8s & 74.8\% / 33.5s & 58.0\% / 115.0s \\
        ARC-Easy & 93.4\% / 33.1s & 92.4\% / 22.0s & 82.0\% / 113.9s & 88.6\% / 51.1s & 92.3\% / 28.2s & 73.7\% / 113.9s \\
        ARC-Challenge & 86.0\% / 49.0s & 82.1\% / 29.5s & 72.4\% / 113.9s & 78.2\% / 68.5s & 81.7\% / 38.2s & 64.0\% / 112.7s \\
        GPQA & 29.8\% / 348.4s & 39.9\% / 174.6s & 28.8\% / 216.2s & 32.3\% / 483.3s & 37.9\% / 225.5s & 28.3\% / 210.1s \\
        MedQA & 53.3\% / 91.5s & 48.0\% / 83.0s & 43.3\% / 137.4s & 44.7\% / 125.0s & 47.0\% / 93.4s & 37.0\% / 133.6s \\
        MBPP-Plus & 50.5\% / 108.7s & 51.3\% / 69.2s & 23.0\% / 132.8s & 37.8\% / 125.6s & 47.9\% / 79.9s & 15.9\% / 131.9s \\
        HumanEval-Plus & 40.9\% / 121.6s & 37.2\% / 80.1s & 15.8\% / 136.4s & 31.1\% / 126.1s & 40.9\% / 100.3s & 7.9\% / 132.6s \\
        AIME 2024 & 23.3\% / 1314.4s & 36.7\% / 385.8s & 3.3\% / 535.6s & 13.3\% / 2806.9s & 23.3\% / 513.3s & 6.7\% / 518.8s \\
        AIME 2025 & 16.7\% / 1432.9s & 26.7\% / 382.0s & 10.0\% / 532.7s & 16.7\% / 1996.1s & 23.3\% / 505.7s & 16.7\% / 519.6s \\
        \midrule
        \textbf{Macro Avg.} & 52.7\% / 391.9s & 54.5\% / 139.2s & 39.1\% / 225.8s & 45.2\% / 649.6s & 52.1\% / 179.8s & 34.2\% / 220.9s \\
        \bottomrule
    \end{tabular}

\end{table}

\section{Rollout-Length Sweep for LatentMAS-Hybrid}
\label{app:latentmas_hybrid_failure}

We vary the rollout length of \textit{LatentMAS-Hybrid}, a public fork of LatentMAS supporting heterogeneous role assignments \citep{zou2025latent,latentmashybrid2026}, and measure task accuracy and answer perplexity.

\begin{figure}[!htbp]
    \centering
    \includegraphics[width=\textwidth]{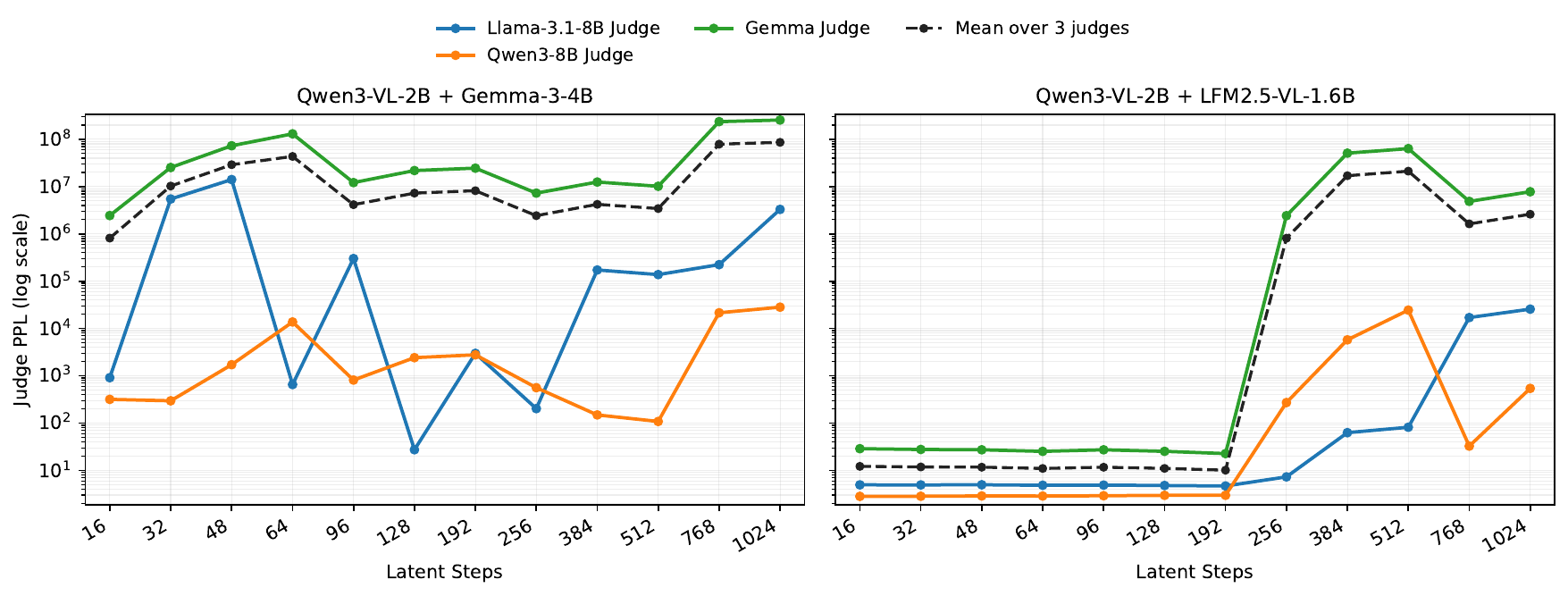}

    \caption{\textbf{Answer perplexity across rollout lengths for the evaluated LatentMAS-Hybrid configurations.}
    Curves show perplexity assigned to the generated answers by the external judge models and their mean (dashed black).}
    \label{fig:latentmas_hybrid_failure}

\end{figure}

\paragraph{Sweep setup.}
We run a rollout-length sweep on GSM8K using a fixed 200-example subset.
We use the public implementation described in Appendix~\ref{app:latent_channels}.
The rollout-length grid is shown in Figure~\ref{fig:latentmas_hybrid_failure}.
The sweep uses greedy decoding, a 256-token answer budget, and batch size 4.
We compare the Qwen/LFM and Qwen/Gemma configurations in Table~\ref{tab:mas_model_configs}.
Table~\ref{tab:hybrid_same_subset} reports representative rollout lengths and VW's main-experiment predictions on these same questions and role assignments.

\begin{table}[!htbp]
    \centering
    \caption{\textbf{LatentMAS-Hybrid and VW on the same GSM8K subset.} Accuracy (\%), with correct counts in parentheses. VW uses the main-run generation settings in Appendix~\ref{app:runtime_budgets}.}
    \label{tab:hybrid_same_subset}
    \small
    \setlength{\tabcolsep}{12pt}
    \begin{tabular}{@{}lrr@{}}
        \toprule
        Method & Gemma / Qwen & LFM / Qwen \\
        \midrule
        LatentMAS-Hybrid, 16 steps & 0.5 & 23.5 \\
        \hspace{1em}48 steps & 0.0 & 26.0 \\
        \hspace{1em}128 steps & 0.0 & 16.5 \\
        \hspace{1em}192 steps & 0.5 & 18.0 \\
        \hspace{1em}256 steps & 0.0 & 20.0 \\
        \hspace{1em}512 steps & 0.0 & 12.5 \\
        \hspace{1em}1024 steps & 0.5 & 14.0 \\
        \midrule
        VW & \textbf{79.0} (158) & \textbf{78.0} (156) \\
        \bottomrule
    \end{tabular}
\end{table}

\paragraph{Metrics.}
At each rollout length, we record GSM8K exact-match accuracy and compute answer perplexity with three language models from the Llama, Qwen, and Gemma families. We also report the mean of their perplexities.

\paragraph{Results.}
For Qwen3-VL-2B + Gemma-3-4B, accuracy never exceeds 0.5\%, and mean judge PPL rises from $8.1\times10^5$ at the shortest evaluated rollout to $8.6\times10^7$ at the longest.
For Qwen3-VL-2B + LFM2.5-VL-1.6B, mean judge PPL is around 10--12 for shorter rollouts, then rises to $8.1\times10^5$ and remains in the $10^6$--$10^7$ range for longer rollouts while accuracy stays around 12--20\%.

\section{Implementations of Prior Latent Communication Channels}
\label{app:latent_channels}

We compare Vision Wormhole (VW) with CIPHER \citep{pham2024cipher} and LatentMAS-Hybrid \citep{latentmashybrid2026} in the four-role Qwen3-VL-2B-Thinking/Gemma-3-4B workflow.
Related interfaces study pairwise activation and cache transfer \citep{ramesh2025activations,fu2025cache}.
\label{app:channel_port}

\paragraph{Setup.}
Evaluation uses the first 256 GSM8K test problems and the full HumanEval-Plus test set. Final-answer caps follow Table~\ref{tab:gen_budgets}. Cross-tokenizer handling follows LatentMAS-Hybrid \citep{latentmashybrid2026}. Table~\ref{tab:channel_port} gives the results and channel settings.

\begin{table}[!ht]
\caption{\textbf{System accuracy for heterogeneous latent communication channels.} Accuracy (\%), with correct counts in parentheses. All methods use the Gemma/Qwen role assignment in Table~\ref{tab:mas_model_configs}.}
\label{tab:channel_port}
\begin{center}
\small
\begin{tabular}{lcc}
\toprule
\textbf{Channel} & \textbf{GSM8K} & \textbf{HumanEval+} \\
\midrule
\textbf{Vision Wormhole} & \textbf{77.0\%} \small(197) & \textbf{37.2\%} \small(61) \\
CIPHER & 65.6\% \small(168) & 21.3\% \small(35) \\
LatentMAS-Hybrid & \phantom{0}0.0\% \small(0) & \phantom{0}0.0\% \small(0) \\
\bottomrule
\end{tabular}
\par\smallskip
\begin{minipage}{\linewidth}
\footnotesize
Channel settings: 256 embedding tokens (CIPHER); 32 latent steps (LatentMAS-Hybrid).
\end{minipage}
\end{center}

\end{table}

Appendix~\ref{app:latentmas_hybrid_failure} reports the LatentMAS-Hybrid rollout-length sweep.

\section{Visual Interface Ablation on Qwen}
\label{app:native_interfaces}

\begin{wraptable}{R}{0.5\textwidth}
    \centering

    \caption{\textbf{Visual interface ablation on Qwen.} Accuracy (\%); differences in pp.}
    \label{tab:native_interfaces}
    \small
    \setlength{\tabcolsep}{4pt}
    \begin{tabular}{@{}lrr@{}}
        \toprule
        Method & GSM8K & MedQA \\
        \midrule
        Qwen + VW & \textbf{75.8} & \textbf{34.4} \\
        Qwen + soft-prefix & 60.2 & 27.0 \\
        $\Delta$ (pp) & +15.6 & +7.4 \\
        \bottomrule
    \end{tabular}
\end{wraptable}

We compare the visual interface with a learned text-prefix interface for Qwen on mathematical and medical reasoning (Table~\ref{tab:native_interfaces}). Both interfaces consume the same intermediate-role trajectories and decode their accumulated messages into a 64-token input span. The visual interface uses the model's native image-token span; the text-prefix interface adds its decoded residual to ordinary space-token embeddings. Each retains its native token structure and attention behavior.

\paragraph{Setup.}
We train separate visual and text-prefix codecs with matched architectures, paired initialization, identical anchor schedules, and the same prompt-boundary hidden-state and KL objectives. The update count and batch size follow Appendix~\ref{app:codec_training_setup}; initialization uses the evaluation seed in Appendix~\ref{app:runtime_budgets}. The RMS regularizer targets each interface's own carrier embeddings. Evaluation uses 256 questions per benchmark, with rollout length $T=T_{\mathrm{train}}$ per intermediate role (Appendix~\ref{app:codec_training_setup}). The two interfaces share the final role instructions and question set; decoding parameters and generation caps follow Appendix~\ref{app:runtime_budgets}. Scoring follows Section~\ref{sec:exp_settings}.

Table~\ref{tab:native_interfaces} shows accuracy gains of 15.6 percentage points on GSM8K and 7.4 points on MedQA with visual reception.

\section{Additional Communication Experiments}
\label{app:communication_experiments}

\subsection{Intermediate Text Budgets}
\label{app:text_budgets}

We vary the maximum generation length of each intermediate TextMAS role in the Gemma/Qwen configuration of Table~\ref{tab:mas_model_configs}. The question sets follow Appendix~\ref{app:latent_channels}; final-answer budgets and sampling settings follow Appendix~\ref{app:runtime_budgets}. All runs use batch size 4. Table~\ref{tab:text_budgets} shows that GSM8K accuracy varies little across these budgets, while HumanEval+ achieves its highest score with the full task budget.

\begin{table}[!ht]
    \centering
    \caption{\textbf{TextMAS with capped intermediate messages.} Accuracy (\%); ``Task cap'' uses the per-task limit in Table~\ref{tab:gen_budgets}.}
    \label{tab:text_budgets}
    \small
    \setlength{\tabcolsep}{14pt}
    \begin{tabular}{@{}lrr@{}}
        \toprule
        Intermediate token cap & GSM8K & HumanEval+ \\
        \midrule
        256 & 79.3 & 34.1 \\
        512 & \textbf{80.5} & --- \\
        1024 & 78.1 & 32.9 \\
        Task cap & \textbf{80.5} & \textbf{40.9} \\
        \bottomrule
    \end{tabular}
\end{table}

\paragraph{Concise intermediate messages.}
VW outperforms both standard and concise TextMAS in five of the six main configurations (Table~\ref{tab:concise_textmas}). We evaluate the first 128 GSM8K test questions per configuration, using the shortest intermediate cap in Table~\ref{tab:text_budgets} and instructing concise TextMAS roles to keep their messages brief. Final-answer generation follows Appendix~\ref{app:runtime_budgets}; standard TextMAS and VW use the corresponding main-experiment predictions.

\begin{table}[!ht]
    \centering
    \caption{\textbf{Standard and concise text communication across all six configurations.} GSM8K accuracy (\%), with correct counts in parentheses. The pooled row aggregates 768 configuration--question evaluations.}
    \label{tab:concise_textmas}
    \small
    \setlength{\tabcolsep}{7pt}
    \begin{tabular}{@{}lrrr@{}}
        \toprule
        Configuration & TextMAS & Concise TextMAS & VW \\
        \midrule
        SmolVLM2 / Gemma & 65.6 (84) & 69.5 (89) & \textbf{82.8} (106) \\
        SmolVLM2 / Qwen & 64.8 (83) & 50.8 (65) & \textbf{72.7} (93) \\
        Gemma / Qwen & \textbf{78.9} (101) & 75.0 (96) & 74.2 (95) \\
        LFM / Gemma & 77.3 (99) & 74.2 (95) & \textbf{85.2} (109) \\
        LFM / Qwen & 68.0 (87) & 64.8 (83) & \textbf{74.2} (95) \\
        Four-model pool & 57.8 (74) & 63.3 (81) & \textbf{72.7} (93) \\
        \midrule
        Pooled & 68.8 (528) & 66.3 (509) & \textbf{77.0} (591) \\
        \bottomrule
    \end{tabular}
\end{table}

\subsection{Parallel Multi-Agent Topology}
\label{app:parallel_topology}

We also evaluate a parallel topology with three independent specialists and a final aggregator, using Gemma-3-4B and Qwen3-VL-8B-Thinking. Gemma supplies two specialists; Qwen supplies the third specialist and the aggregator. Aggregation instructions are adapted to each communication channel, and VW reuses the trained codec. Table~\ref{tab:parallel_topology} reports accuracy on a prefix of the GSM8K test set and the full HumanEval+ test set, with sample counts in the table and final-answer caps from Table~\ref{tab:gen_budgets}. VW achieves the highest accuracy on both tasks.

\begin{table}[!htbp]
    \centering
    \caption{\textbf{Communication channels in a parallel MAS.} Accuracy (\%), with correct counts in parentheses.}
    \label{tab:parallel_topology}
    \small
    \setlength{\tabcolsep}{12pt}
    \begin{tabular}{@{}lrrrr@{}}
        \toprule
        Task & $n$ & Text & OCR relay & VW \\
        \midrule
        GSM8K & 128 & 90.6 (116) & 92.2 (118) & \textbf{96.9} (124) \\
        HumanEval+ & 164 & 64.0 (105) & 44.5 (73) & \textbf{65.2} (107) \\
        \bottomrule
    \end{tabular}
\end{table}

\subsection{Cross-Model Key--Value Retrieval}
\label{app:cross_model_retrieval}

We evaluate Qwen-to-Gemma retrieval from key--value memories built from GSM8K question--answer pairs (Table~\ref{tab:cross_model_retrieval}). Qwen reads the target pair and 16 distractors. Gemma receives the message and query question, then selects the answer from shuffled candidate values. VW uses the frozen main-experiment codec and alignment with rollout length $T=T_{\mathrm{train}}$ (Appendix~\ref{app:codec_training_setup}); text communication uses a sender-generated memory note. We evaluate 100 queries per seed over three seeds, with shared queries, candidates, receiver decoding, and a 256-token answer cap.

\begin{table}[!ht]
    \centering
    \caption{\textbf{Key--value retrieval through the Qwen $\to$ Gemma interface.} Exact-selection accuracy (\%), reported as mean $\pm$ standard deviation across seeds.}
    \label{tab:cross_model_retrieval}
    \small
    \setlength{\tabcolsep}{14pt}
    \begin{tabular}{@{}lr@{}}
        \toprule
        Communication & Retrieval accuracy \\
        \midrule
        Text message & $34.0 \pm 8.7$ \\
        Vision Wormhole & $\mathbf{36.0} \pm 4.6$ \\
        \bottomrule
    \end{tabular}
\end{table}

\paragraph{Retrieval with larger memories.}
VW retrieves more accurately than text communication in larger memories (Table~\ref{tab:gemma_memory_retrieval}). We vary the number of distractors for Gemma-to-Gemma communication using its frozen codec, with 30 queries per seed for each condition over the same seeds. Text senders use the shortest intermediate cap in Table~\ref{tab:text_budgets}.

\begin{table}[!ht]
    \centering
    \caption{\textbf{Gemma retrieval as memory size increases.} Exact-selection accuracy (\%), reported as mean $\pm$ standard deviation across seeds.}
    \label{tab:gemma_memory_retrieval}
    \small
    \setlength{\tabcolsep}{10pt}
    \begin{tabular}{@{}rrr@{}}
        \toprule
        Distractor entries & Text message & Vision Wormhole \\
        \midrule
        0 & $\mathbf{100.0} \pm 0.0$ & $88.9 \pm 5.7$ \\
        1 & $\mathbf{91.1} \pm 4.2$ & $64.4 \pm 7.9$ \\
        4 & $\mathbf{66.7} \pm 7.2$ & $53.3 \pm 7.2$ \\
        16 & $17.8 \pm 9.6$ & $\mathbf{34.4} \pm 6.8$ \\
        64 & $3.3 \pm 4.7$ & $\mathbf{26.7} \pm 9.8$ \\
        \bottomrule
    \end{tabular}
\end{table}

\subsection{Representation Alignment}
\label{app:heldout_reconstruction}

Aligned messages achieve lower reconstruction error than cross-text substitutions and shuffled-anchor maps in both directions (Table~\ref{tab:heldout_reconstruction}). We train Qwen and Gemma ablation codecs using the visual-interface recipe in Appendix~\ref{app:native_interfaces} and fit affine maps on paired calibration anchors. Evaluation uses 354 held-out texts across three training seeds, measuring prompt-boundary KL divergence from the receiver's text-conditioned teacher to its message-conditioned output. The controls substitute messages across texts or shuffle the anchor correspondences when fitting the maps.

\begin{table}[!ht]
    \centering
    \caption{\textbf{Aligned messages reduce reconstruction error.} Mean KL divergence ($\downarrow$) on held-out texts.}
    \label{tab:heldout_reconstruction}
    \small
    \setlength{\tabcolsep}{9pt}
    \begin{tabular}{@{}lrrr@{}}
        \toprule
        Direction & Aligned message & Cross-text message & Shuffled-anchor map \\
        \midrule
        Gemma $\to$ Qwen & \textbf{0.227} & 0.513 & 0.464 \\
        Qwen $\to$ Gemma & \textbf{1.864} & 6.344 & 5.408 \\
        \bottomrule
    \end{tabular}
\end{table}

\paragraph{Affine calibration of pooled representations.}
We additionally examine a larger Qwen/Gemma codec with 1024-dimensional messages on 1,000 Flickr30K captions. A ridge-affine readout is fitted to the paired mean-pooled representations, with ridge coefficient 0.1. Table~\ref{tab:affine_calibration} reports calibration-set centered kernel alignment (CKA) and caption-identity retrieval, averaged over both directions. Calibration increases representation agreement and cross-model retrieval accuracy; shuffling caption correspondences after calibration gives CKA 0.003.

\begin{table}[!htbp]
    \centering
    \caption{\textbf{Affine calibration of cross-model representations.} CKA and retrieval accuracy (\%) on the paired calibration captions.}
    \label{tab:affine_calibration}
    \small
    \setlength{\tabcolsep}{12pt}
    \begin{tabular}{@{}lrrrr@{}}
        \toprule
        Readout & CKA & Top-1 & Top-5 & Top-10 \\
        \midrule
        Raw pooled representations & 0.261 & 0.20 & 1.15 & 2.45 \\
        Affine-calibrated & \textbf{0.840} & \textbf{10.05} & \textbf{17.15} & \textbf{22.15} \\
        Chance retrieval & --- & 0.10 & 0.50 & 1.00 \\
        \bottomrule
    \end{tabular}
\end{table}

\subsection{Input-Dependent Decoder Gates}
\label{app:gate_response}

The frozen Qwen decoder assigns substantially larger gates to caption-derived messages than to Gaussian inputs (Table~\ref{tab:gate_response}). We compare messages produced from 20 Flickr30K captions with 200 Gaussian message tensors of the same shape, sampled from $\mathcal{N}(0,0.5^2)$. The decoder weights are shared across both conditions.

\begin{table}[!htbp]
    \centering
    \caption{\textbf{Qwen decoder gate response.} Mean gate value and standard deviation across inputs.}
    \label{tab:gate_response}
    \small
    \setlength{\tabcolsep}{18pt}
    \begin{tabular}{@{}lr@{}}
        \toprule
        Decoder input & Gate value \\
        \midrule
        Gaussian message & $0.0185 \pm 0.0001$ \\
        Caption-derived message & $0.6551 \pm 0.0066$ \\
        \bottomrule
    \end{tabular}
\end{table}

\subsection{Completeness of Intermediate Role Outputs}
\label{app:sender_quality}

We examine intermediate outputs in the sequential Gemma-3-4B/Qwen3-VL-8B TextMAS configuration. Claude Opus grades each planner, critic, and refiner output against the reference answer as correct, partial, incorrect, or no attempt. Table~\ref{tab:sender_quality} counts roles graded fully correct for each question. The prevalence of partial solutions motivates preserving intermediate reasoning across roles.

\begin{table}[!htbp]
    \centering
    \caption{\textbf{Completeness of sender outputs.} Percentage of questions with the indicated number of fully correct intermediate roles.}
    \label{tab:sender_quality}
    \small
    \setlength{\tabcolsep}{12pt}
    \begin{tabular}{@{}lrrr@{}}
        \toprule
        Fully correct senders & GSM8K ($n=100$) & HumanEval+ ($n=100$) & AIME 2024 ($n=30$) \\
        \midrule
        Three & 39.0 & 8.0 & 0.0 \\
        Two & 45.0 & 24.0 & 13.3 \\
        One & 12.0 & 53.0 & 13.3 \\
        Zero & 4.0 & 15.0 & 73.3 \\
        \bottomrule
    \end{tabular}
\end{table}


\begin{thebibliography}{156}
\providecommand{\natexlab}[1]{#1}
\providecommand{\url}[1]{\texttt{#1}}
\expandafter\ifx\csname urlstyle\endcsname\relax
  \providecommand{\doi}[1]{doi: #1}\else
  \providecommand{\doi}{doi: \begingroup \urlstyle{rm}\Url}\fi

\bibitem[Acharya et~al.(2025)Acharya, Kuppan, and
  Bhaskaracharya]{acharya2025agentic}
Deepak~Bhaskar Acharya, Karthigeyan Kuppan, and Divya Bhaskaracharya.
\newblock Agentic ai: Autonomous intelligence for complex goals -- a
  comprehensive survey.
\newblock \emph{IEEE Access}, 13:\penalty0 18912--18936, 2025.
\newblock \doi{10.1109/ACCESS.2025.3532853}.

\bibitem[Ainsworth et~al.(2022)Ainsworth, Hayase, and
  Srinivasa]{ainsworth2022git}
Samuel~K Ainsworth, Jonathan Hayase, and Siddhartha Srinivasa.
\newblock Git re-basin: Merging models modulo permutation symmetries.
\newblock \emph{arXiv preprint arXiv:2209.04836}, 2022.

\bibitem[Alayrac et~al.(2022)Alayrac, Donahue, Luc, Miech, Barr, Hasson, Lenc,
  Mensch, Millican, Reynolds, Ring, Rutherford, Cabi, Han, Gong, Samangooei,
  Monteiro, Menick, Borgeaud, Brock, Nematzadeh, Sharifzadeh, Binkowski,
  Barreira, Vinyals, Zisserman, and Simonyan]{alayrac2022flamingo}
Jean-Baptiste Alayrac, Jeff Donahue, Pauline Luc, Antoine Miech, Iain Barr,
  Yana Hasson, Karel Lenc, Arthur Mensch, Katie Millican, Malcolm Reynolds,
  Roman Ring, Eliza Rutherford, Serkan Cabi, Tengda Han, Zhitao Gong, Sina
  Samangooei, Marianne Monteiro, Jacob Menick, Sebastian Borgeaud, Andrew
  Brock, Aida Nematzadeh, Sahand Sharifzadeh, Mikolaj Binkowski, Ricardo
  Barreira, Oriol Vinyals, Andrew Zisserman, and Karen Simonyan.
\newblock Flamingo: a visual language model for few-shot learning.
\newblock In \emph{NeurIPS}, 2022.

\bibitem[Amini et~al.(2025)Amini, Banaszak, Benoit, B{\"o}{\"o}k, Dakhran,
  Duong, Eng, Fernandes, H{\"a}rk{\"o}nen, Harrington, Hasani, Karwa,
  Khrustalev, Labonne, Lechner, Lechner, Lee, Li, Loo, Marks, Mosca, Paech,
  Pak, Parnichkun, Quach, Rogers, Rus, Saxena, Schlager, Seyde, Smith,
  Tadimeti, and Tumma]{amini2025lfm2technicalreport}
Alexander Amini, Anna Banaszak, Harold Benoit, Arthur B{\"o}{\"o}k, Tarek
  Dakhran, Song Duong, Alfred Eng, Fernando Fernandes, Marc H{\"a}rk{\"o}nen,
  Anne Harrington, Ramin Hasani, Saniya Karwa, Yuri Khrustalev, Maxime Labonne,
  Mathias Lechner, Valentine Lechner, Simon Lee, Zetian Li, Noel Loo, Jacob
  Marks, Edoardo Mosca, Samuel~J. Paech, Paul Pak, Rom~N. Parnichkun, Alex
  Quach, Ryan Rogers, Daniela Rus, Nayan Saxena, Bettina Schlager, Tim Seyde,
  Jimmy T.~H. Smith, Aditya Tadimeti, and Neehal Tumma.
\newblock Lfm2 technical report, 2025.
\newblock URL \url{https://arxiv.org/abs/2511.23404}.

\bibitem[Anantha et~al.(2023)Anantha, Bandyopadhyay, Kashi, Mahinder, Hill, and
  Chappidi]{anantha2023protipprogressivetoolretrieval}
Raviteja Anantha, Bortik Bandyopadhyay, Anirudh Kashi, Sayantan Mahinder,
  Andrew~W Hill, and Srinivas Chappidi.
\newblock Protip: Progressive tool retrieval improves planning, 2023.
\newblock URL \url{https://arxiv.org/abs/2312.10332}.

\bibitem[Asif et~al.(2026)Asif, Amiri, Abbas, Sattigeri, and
  Ramamurthy]{asif2026lcguardlatentcommunicationguard}
Sadia Asif, Mohammad~Mohammadi Amiri, Momin Abbas, Prasanna Sattigeri, and
  Karthikeyan~Natesan Ramamurthy.
\newblock Lcguard: Latent communication guard for safe kv sharing in
  multi-agent systems, 2026.
\newblock URL \url{https://arxiv.org/abs/2605.22786}.

\bibitem[Bai et~al.(2025)Bai, Cai, Chen, Chen, Chen, Cheng, Deng, Ding, Gao,
  Ge, Ge, Guo, Huang, Huang, Huang, Hui, Jiang, Li, Li, Li, Li, Lin, Lin, Liu,
  Liu, Liu, Liu, Liu, Liu, Lu, Luo, Lv, Men, Meng, Ren, Ren, Song, Sun, Tang,
  Tu, Wan, Wang, Wang, Wang, Wang, Xie, Xu, Xu, Xu, Yang, Yang, Yang, Yang, Yu,
  Zhang, Zhang, Zhang, Zheng, Zhong, Zhou, Zhou, Zhou, Zhu, and
  Zhu]{bai2025qwen3vltechnicalreport}
Shuai Bai, Yuxuan Cai, Ruizhe Chen, Keqin Chen, Xionghui Chen, Zesen Cheng,
  Lianghao Deng, Wei Ding, Chang Gao, Chunjiang Ge, Wenbin Ge, Zhifang Guo,
  Qidong Huang, Jie Huang, Fei Huang, Binyuan Hui, Shutong Jiang, Zhaohai Li,
  Mingsheng Li, Mei Li, Kaixin Li, Zicheng Lin, Junyang Lin, Xuejing Liu,
  Jiawei Liu, Chenglong Liu, Yang Liu, Dayiheng Liu, Shixuan Liu, Dunjie Lu,
  Ruilin Luo, Chenxu Lv, Rui Men, Lingchen Meng, Xuancheng Ren, Xingzhang Ren,
  Sibo Song, Yuchong Sun, Jun Tang, Jianhong Tu, Jianqiang Wan, Peng Wang,
  Pengfei Wang, Qiuyue Wang, Yuxuan Wang, Tianbao Xie, Yiheng Xu, Haiyang Xu,
  Jin Xu, Zhibo Yang, Mingkun Yang, Jianxin Yang, An~Yang, Bowen Yu, Fei Zhang,
  Hang Zhang, Xi~Zhang, Bo~Zheng, Humen Zhong, Jingren Zhou, Fan Zhou, Jing
  Zhou, Yuanzhi Zhu, and Ke~Zhu.
\newblock Qwen3-vl technical report, 2025.
\newblock URL \url{https://arxiv.org/abs/2511.21631}.

\bibitem[Bansal et~al.(2024)Bansal, Samanta, Dalmia, Gupta, Vashishth,
  Ganapathy, Bapna, Jain, and Talukdar]{bansal2024llmaugmentedllmsexpanding}
Rachit Bansal, Bidisha Samanta, Siddharth Dalmia, Nitish Gupta, Shikhar
  Vashishth, Sriram Ganapathy, Abhishek Bapna, Prateek Jain, and Partha
  Talukdar.
\newblock Llm augmented llms: Expanding capabilities through composition, 2024.
\newblock URL \url{https://arxiv.org/abs/2401.02412}.

\bibitem[Bansal et~al.(2021)Bansal, Nakkiran, and Barak]{bansal2021revisiting}
Yamini Bansal, Preetum Nakkiran, and Boaz Barak.
\newblock Revisiting model stitching to compare neural representations.
\newblock In \emph{Advances in Neural Information Processing Systems 34
  (NeurIPS 2021)}, 2021.
\newblock URL
  \url{https://proceedings.neurips.cc/paper/2021/file/01ded4259d101feb739b06c399e9cd9c-Paper.pdf}.

\bibitem[Berrayana et~al.(2026)Berrayana, Heakl, Sohail, Hofmann, Khan, and
  Chen]{berrayana2026latentdarmbridgingdiscretediffusion}
Lina Berrayana, Ahmed Heakl, Abdullah Sohail, Thomas Hofmann, Salman Khan, and
  Wei Chen.
\newblock Latent-darm: Bridging discrete diffusion and autoregressive models
  for reasoning, 2026.
\newblock URL \url{https://arxiv.org/abs/2603.09184}.

\bibitem[Bian et~al.(2026)Bian, Wu, Zhang, Dong, Liang, and
  Zhuo]{bian2026tokendancescalingmultiagentllm}
Zhuohang Bian, Feiyang Wu, Chengrui Zhang, Hangcheng Dong, Yun Liang, and
  Youwei Zhuo.
\newblock Tokendance: Scaling multi-agent llm serving via collective kv cache
  sharing, 2026.
\newblock URL \url{https://arxiv.org/abs/2604.03143}.

\bibitem[Birr et~al.(2024)Birr, Pohl, Younes, and
  Asfour]{birr2024autogptpaffordancebasedtaskplanning}
Timo Birr, Christoph Pohl, Abdelrahman Younes, and Tamim Asfour.
\newblock Autogpt+p: Affordance-based task planning with large language models,
  2024.
\newblock URL \url{https://arxiv.org/abs/2402.10778}.

\bibitem[Brito \& Baquero(2026)Brito and
  Baquero]{brito2026latentagentsliekvcache}
Luís Brito and Carlos Baquero.
\newblock When latent agents lie: Kv-cache integrity in multi-agent llm
  collaboration, 2026.
\newblock URL \url{https://arxiv.org/abs/2606.28958}.

\bibitem[Cemri et~al.(2025)Cemri, Pan, Yang, Agrawal, Chopra, Tiwari, Keutzer,
  Parameswaran, Klein, Ramchandran, Zaharia, Gonzalez, and
  Stoica]{cemri2025multi}
Mert Cemri, Melissa~Z. Pan, Shuyi Yang, Lakshya~A. Agrawal, Bhavya Chopra,
  Rishabh Tiwari, Kurt Keutzer, Aditya~G. Parameswaran, Dan Klein, Kannan
  Ramchandran, Matei Zaharia, Joseph~E. Gonzalez, and Ion Stoica.
\newblock Why do multi-agent llm systems fail?
\newblock \emph{arXiv preprint arXiv:2503.13657}, 2025.

\bibitem[Chen et~al.(2026{\natexlab{a}})Chen, Yang, Liu, and
  Joe-Wong]{chen2026wsmultiagentcommunicationtalks}
Jingdi Chen, Hanqing Yang, Zongjun Liu, and Carlee Joe-Wong.
\newblock The five ws of multi-agent communication: Who talks to whom, when,
  what, and why -- a survey from marl to emergent language and llms,
  2026{\natexlab{a}}.
\newblock URL \url{https://arxiv.org/abs/2602.11583}.

\bibitem[Chen et~al.(2026{\natexlab{b}})Chen, Wang, Zhang, Li, Lu, and
  Fan]{chen2026scalingvideounderstandingcompact}
Kerui Chen, Jinglu Wang, Jianrong Zhang, Ming Li, Yan Lu, and Hehe Fan.
\newblock Scaling video understanding via compact latent multi-agent
  collaboration, 2026{\natexlab{b}}.
\newblock URL \url{https://arxiv.org/abs/2605.00444}.

\bibitem[Chen et~al.(2026{\natexlab{c}})Chen, Zhang, Wu, Tremblay, Blukis,
  Birchfield, Vidal, Velasquez, Liu, and Qu]{chen2026iseeknowi}
Siyi Chen, Xiaoyan Zhang, Meng Wu, Jonathan Tremblay, Valts Blukis, Stan
  Birchfield, Rene Vidal, Alvaro Velasquez, Sijia Liu, and Qing Qu.
\newblock See what i see, know what i think: Dense latent communication across
  heterogeneous agents, 2026{\natexlab{c}}.
\newblock URL \url{https://arxiv.org/abs/2606.13594}.

\bibitem[Chen et~al.(2026{\natexlab{d}})Chen, Wu, and
  Lee]{chen2026lacoadaptivelatentcommunication}
Tianhao Chen, Yuheng Wu, and Dongman Lee.
\newblock Laco: Adaptive latent communication for collaborative driving,
  2026{\natexlab{d}}.
\newblock URL \url{https://arxiv.org/abs/2605.22504}.

\bibitem[Chen et~al.(2024)Chen, Yuan, Qian, Yang, Liu, and Sun]{chen2024optima}
Weize Chen, Jiarui Yuan, Chen Qian, Cheng Yang, Zhiyuan Liu, and Maosong Sun.
\newblock Optima: Optimizing effectiveness and efficiency for llm-based
  multi-agent system.
\newblock \emph{arXiv preprint arXiv:2410.08115}, 2024.

\bibitem[Chen et~al.(2025)Chen, Yuan, Qian, Yang, Liu, and Sun]{chen2025optima}
Weize Chen, Jiarui Yuan, Chen Qian, Cheng Yang, Zhiyuan Liu, and Maosong Sun.
\newblock Optima: Optimizing effectiveness and efficiency for llm-based
  multi-agent system.
\newblock In \emph{Findings of the Association for Computational Linguistics:
  ACL 2025}, pp.\  11534--11557, 2025.

\bibitem[Cheng et~al.(2026)Cheng, Das, and
  Ramnath]{cheng2026doeslatentcommunicationpay}
Jiaming Cheng, Subhransu Das, and Rajiv Ramnath.
\newblock When does latent communication pay? a causal audit of relayed kv
  caches in multi-agent llms, 2026.
\newblock URL \url{https://arxiv.org/abs/2608.04893}.

\bibitem[Clark et~al.(2018{\natexlab{a}})Clark, Cowhey, Etzioni, Khot,
  Sabharwal, Schoenick, and Tafjord]{arc-challenge}
Peter Clark, Isaac Cowhey, Oren Etzioni, Tushar Khot, Ashish Sabharwal, Carissa
  Schoenick, and Oyvind Tafjord.
\newblock Think you have solved question answering? try arc, the ai2 reasoning
  challenge.
\newblock \emph{arXiv preprint arXiv:1803.05457}, 2018{\natexlab{a}}.

\bibitem[Clark et~al.(2018{\natexlab{b}})Clark, Cowhey, Etzioni, Khot,
  Sabharwal, Schoenick, and Tafjord]{arc-easy}
Peter Clark, Isaac Cowhey, Oren Etzioni, Tushar Khot, Ashish Sabharwal, Carissa
  Schoenick, and Oyvind Tafjord.
\newblock Think you have solved question answering? try arc, the ai2 reasoning
  challenge.
\newblock \emph{arXiv preprint arXiv:1803.05457}, 2018{\natexlab{b}}.

\bibitem[Cobbe et~al.(2021)Cobbe, Kosaraju, Bavarian, Chen, Jun, Kaiser,
  Plappert, Tworek, Hilton, Nakano, Hesse, and Schulman]{gsm8k}
Karl Cobbe, Vineet Kosaraju, Mohammad Bavarian, Mark Chen, Heewoo Jun, Lukasz
  Kaiser, Matthias Plappert, Jerry Tworek, Jacob Hilton, Reiichiro Nakano,
  Christopher Hesse, and John Schulman.
\newblock Training verifiers to solve math word problems, 2021.
\newblock URL \url{https://arxiv.org/abs/2110.14168}.

\bibitem[Coda-Forno et~al.(2025)Coda-Forno, Zhao, Zhang, Tamboli, Li, Fan,
  Zhang, Schulz, and Tseng]{coda2025exploring}
Julian Coda-Forno, Zhuokai Zhao, Qiang Zhang, Dipesh Tamboli, Weiwei Li,
  Xiangjun Fan, Lizhu Zhang, Eric Schulz, and Hsiao-Ping Tseng.
\newblock Exploring system 1 and 2 communication for latent reasoning in llms.
\newblock \emph{arXiv preprint arXiv:2510.00494}, 2025.

\bibitem[Dai et~al.(2026)Dai, Xiang, and
  Yang]{dai2026tokenkvcachecommunicationmediaselection}
Lipeng Dai, Luping Xiang, and Kun Yang.
\newblock A token/kv-cache communication media selection and resource
  allocation strategy for multi-agent collaboration, 2026.
\newblock URL \url{https://arxiv.org/abs/2605.25422}.

\bibitem[Downey et~al.(2023)Downey, Blevins, Goldfine, and
  Steinert-Threlkeld]{downey2023embedding}
C.m. Downey, Terra Blevins, Nora Goldfine, and Shane Steinert-Threlkeld.
\newblock Embedding structure matters: Comparing methods to adapt multilingual
  vocabularies to new languages.
\newblock In Duygu Ataman (ed.), \emph{Proceedings of the 3rd Workshop on
  Multi-lingual Representation Learning (MRL)}, pp.\  268--281, Singapore, dec
  2023. Association for Computational Linguistics.
\newblock \doi{10.18653/v1/2023.mrl-1.20}.
\newblock URL \url{https://aclanthology.org/2023.mrl-1.20/}.

\bibitem[Du et~al.(2025)Du, Huang, Zheng, Wang, Montella, Lapata, Wong, and
  Pan]{du2025rethinking}
Yiming Du, Wenyu Huang, Danna Zheng, Zhaowei Wang, Sebastien Montella, Mirella
  Lapata, Kam-Fai Wong, and Jeff~Z. Pan.
\newblock Rethinking memory in llm based agents: Representations, operations,
  and emerging topics.
\newblock \emph{arXiv preprint arXiv:2505.00675}, 2025.
\newblock \doi{10.48550/arXiv.2505.00675}.

\bibitem[Du et~al.(2026)Du, Wang, Bai, Cao, Zhu, Cheng, Zheng, Chen, and
  Ying]{du-etal-2026-enabling}
Zhuoyun Du, Runze Wang, Huiyu Bai, Zouying Cao, Xiaoyong Zhu, Yu~Cheng,
  Bo~Zheng, Wei Chen, and Haochao Ying.
\newblock Enabling agents to communicate entirely in latent space.
\newblock In Maria Liakata, Viviane~P. Moreira, Jiajun Zhang, and David Jurgens
  (eds.), \emph{Proceedings of the 64th Annual Meeting of the {A}ssociation for
  {C}omputational {L}inguistics (Volume 1: Long Papers)}, pp.\  27106--27129,
  San Diego, California, United States, July 2026. Association for
  Computational Linguistics.
\newblock ISBN 979-8-89176-390-6.
\newblock \doi{10.18653/v1/2026.acl-long.1248}.
\newblock URL \url{https://aclanthology.org/2026.acl-long.1248/}.

\bibitem[Feher et~al.(2025)Feher, Vuli{\'c}, and
  Minixhofer]{feher2025retrofitting}
Darius Feher, Ivan Vuli{\'c}, and Benjamin Minixhofer.
\newblock Retrofitting large language models with dynamic tokenization.
\newblock In \emph{Proceedings of the 63rd Annual Meeting of the Association
  for Computational Linguistics}, 2025.

\bibitem[Fein-Ashley \& Fein-Ashley(2025)Fein-Ashley and
  Fein-Ashley]{fein2025bridging}
Benjamin Fein-Ashley and Jacob Fein-Ashley.
\newblock Bridging hidden states in vision-language models.
\newblock \emph{arXiv preprint arXiv:2511.11526}, 2025.

\bibitem[Fein-Ashley et~al.(2025)Fein-Ashley, Parikh, Kannan, and
  Prasanna]{fein2025mixture}
Jacob Fein-Ashley, Dhruv Parikh, Rajgopal Kannan, and Viktor Prasanna.
\newblock Mixture of thoughts: Learning to aggregate what experts think, not
  just what they say.
\newblock \emph{arXiv preprint arXiv:2509.21164}, 2025.

\bibitem[Feng et~al.(2025)Feng, Xue, Yuan, Yu, Ding, Liu, Gao, Sun, Zheng, and
  Wang]{feng2025multi}
Zhaohan Feng, Ruiqi Xue, Lei Yuan, Yang Yu, Ning Ding, Meiqin Liu, Bingzhao
  Gao, Jian Sun, Xinhu Zheng, and Gang Wang.
\newblock Multi-agent embodied ai: Advances and future directions.
\newblock \emph{arXiv preprint arXiv:2505.05108}, 2025.

\bibitem[Fourney et~al.(2024)Fourney, Bansal, Mozannar, Tan, Salinas, Zhu,
  Niedtner, Proebsting, Bassman, Gerrits, Alber, Chang, Loynd, West, Dibia,
  Awadallah, Kamar, Hosn, and Amershi]{fourney2024magentic}
Adam Fourney, Gagan Bansal, Hussein Mozannar, Cheng Tan, Eduardo Salinas,
  Erkang~(Eric) Zhu, Friederike Niedtner, Grace Proebsting, Griffin Bassman,
  Jack Gerrits, Jacob Alber, Peter Chang, Ricky Loynd, Robert West, Victor
  Dibia, Ahmed Awadallah, Ece Kamar, Rafah Hosn, and Saleema Amershi.
\newblock Magentic-one: A generalist multi-agent system for solving complex
  tasks.
\newblock \emph{arXiv preprint arXiv:2411.04468}, 2024.
\newblock URL \url{https://arxiv.org/abs/2411.04468}.
\newblock Submitted Nov 7, 2024.

\bibitem[Fu et~al.(2025)Fu, Min, Zhang, Yan, Dai, Ouyang, and
  Wang]{fu2025cache}
Tianyu Fu, Zihan Min, Hanling Zhang, Jichao Yan, Guohao Dai, Wanli Ouyang, and
  Yu~Wang.
\newblock Cache-to-cache: Direct semantic communication between large language
  models.
\newblock \emph{arXiv preprint arXiv:2510.03215}, 2025.

\bibitem[Geng et~al.(2026)Geng, Gao, Wu, Liu, and
  Liu]{geng2026relaycachingacceleratingllmcollaboration}
Yingsheng Geng, Yuchong Gao, Weihong Wu, Guyue Liu, and Jiang Liu.
\newblock Relaycaching: Accelerating llm collaboration via decoding kv cache
  reuse, 2026.
\newblock URL \url{https://arxiv.org/abs/2603.13289}.

\bibitem[Goddard \& Neto(2025)Goddard and Neto]{goddard2025training}
C~Goddard and FF~Neto.
\newblock Training-free tokenizer transplantation via orthogonal matching
  pursuit.
\newblock \emph{arXiv preprint arXiv:2506.06607}, 2025.

\bibitem[Guo et~al.(2024)Guo, Chen, Wang, Chang, Pei, Chawla, Wiest, and
  Zhang]{guo2024large}
Taicheng Guo, Xiuying Chen, Yaqi Wang, Ruidi Chang, Shichao Pei, Nitesh~V
  Chawla, Olaf Wiest, and Xiangliang Zhang.
\newblock Large language model based multi-agents: A survey of progress and
  challenges.
\newblock \emph{arXiv preprint arXiv:2402.01680}, 2024.

\bibitem[Hao et~al.(2026)Hao, Dai, Qin, and
  Yu]{hao2026braininspiredgraphmultiagentsystems}
Guangfu Hao, Yuming Dai, Xianzhe Qin, and Shan Yu.
\newblock Brain-inspired graph multi-agent systems for llm reasoning, 2026.
\newblock URL \url{https://arxiv.org/abs/2603.15371}.

\bibitem[Hao et~al.(2024)Hao, Sukhbaatar, Su, Li, Hu, Weston, and
  Tian]{hao2024training}
Shibo Hao, Sainbayar Sukhbaatar, DiJia Su, Xian Li, Zhiting Hu, Jason Weston,
  and Yuandong Tian.
\newblock Training large language models to reason in a continuous latent
  space.
\newblock \emph{arXiv preprint arXiv:2412.06769}, 2024.

\bibitem[Hong et~al.(2024)Hong, Zhuge, Chen, Zheng, Cheng, Wang, Zhang, Wang,
  Yau, Lin, Zhou, Ran, Xiao, Wu, and Schmidhuber]{hong2023metagpt}
Sirui Hong, Mingchen Zhuge, Jonathan Chen, Xiawu Zheng, Yuheng Cheng, Jinlin
  Wang, Ceyao Zhang, Zili Wang, Steven Ka~Shing Yau, Zijuan Lin, Liyang Zhou,
  Chenyu Ran, Lingfeng Xiao, Chenglin Wu, and J\"urgen Schmidhuber.
\newblock Metagpt: Meta programming for a multi-agent collaborative framework.
\newblock In \emph{International Conference on Learning Representations
  (ICLR)}, 2024.

\bibitem[Hou et~al.(2025)Hou, Tang, and Wang]{hou2025halo}
Zhipeng Hou, Junyi Tang, and Yipeng Wang.
\newblock Halo: Hierarchical autonomous logic-oriented orchestration for
  multi-agent llm systems.
\newblock \emph{arXiv preprint arXiv:2505.13516}, 2025.

\bibitem[Hu et~al.(2025{\natexlab{a}})Hu, Chen, Chen, Mu, Shao, and
  Luo]{hu2025hiagent}
Mengkang Hu, Tianxing Chen, Qiguang Chen, Yao Mu, Wenqi Shao, and Ping Luo.
\newblock Hiagent: Hierarchical working memory management for solving
  long-horizon agent tasks with large language model.
\newblock In \emph{Proceedings of the 63rd Annual Meeting of the Association
  for Computational Linguistics (Volume 1: Long Papers)}, pp.\  32779--32798,
  2025{\natexlab{a}}.

\bibitem[Hu et~al.(2025{\natexlab{b}})Hu, Zhou, Fan, Nie, Xia, Sun, Ye, Jin,
  Li, Chen, Zhang, Wang, Ye, Ghanem, Luo, and Li]{hu2025owl}
Mengkang Hu, Yuhang Zhou, Wendong Fan, Yuzhou Nie, Bowei Xia, Tao Sun, Ziyu Ye,
  Zhaoxuan Jin, Yingru Li, Qiguang Chen, Zeyu Zhang, Yifeng Wang, Qianshuo Ye,
  Bernard Ghanem, Ping Luo, and Guohao Li.
\newblock Owl: Optimized workforce learning for general multi-agent assistance
  in real-world task automation.
\newblock \emph{arXiv preprint arXiv:2505.23885}, 2025{\natexlab{b}}.

\bibitem[Jeon et~al.(2026)Jeon, Ha, and Kim]{jeon2026lragentefficientkvcache}
Hyesung Jeon, Hyeongju Ha, and Jae-Joon Kim.
\newblock Lragent: Efficient kv cache sharing for multi-lora llm agents, 2026.
\newblock URL \url{https://arxiv.org/abs/2602.01053}.

\bibitem[Jia et~al.(2021)Jia, Yang, Xia, Chen, Parekh, Pham, Le, Sung, Li, and
  Duerig]{jia2021align}
Chao Jia, Yinfei Yang, Ye~Xia, Yi-Ting Chen, Zarana Parekh, Hieu Pham, Quoc~V.
  Le, Yun-Hsuan Sung, Zhen Li, and Tom Duerig.
\newblock Scaling up visual and vision-language representation learning with
  noisy text supervision.
\newblock In \emph{Proceedings of the 38th International Conference on Machine
  Learning}, volume 139 of \emph{Proceedings of Machine Learning Research},
  pp.\  4904--4916. PMLR, 2021.

\bibitem[Jin et~al.(2026)Jin, Kuang, Yu, Yuan, and
  Wang]{jin2026agentprimitivesreusablelatent}
Haibo Jin, Peng Kuang, Ye~Yu, Xiaopeng Yuan, and Haohan Wang.
\newblock Agent primitives: Reusable latent building blocks for multi-agent
  systems, 2026.
\newblock URL \url{https://arxiv.org/abs/2602.03695}.

\bibitem[Kriuk \& Ng(2025)Kriuk and
  Ng]{kriuk2025qkvcommefficientmultiagentcommunication}
Boris Kriuk and Logic Ng.
\newblock Q-kvcomm: Efficient multi-agent communication via adaptive kv cache
  compression, 2025.
\newblock URL \url{https://arxiv.org/abs/2512.17914}.

\bibitem[Lee et~al.(2026)Lee, Park, Zheng, and
  Park]{lee2026lowlatencyedgellmhandover}
Seunghun Lee, Jihong Park, Ce~Zheng, and Hyuncheol Park.
\newblock Low-latency edge llm handover via joint kv cache transfer and token
  prefill, 2026.
\newblock URL \url{https://arxiv.org/abs/2603.28018}.

\bibitem[Lenc \& Vedaldi(2015)Lenc and Vedaldi]{lenc2015equivariance}
Karel Lenc and Andrea Vedaldi.
\newblock Understanding image representations by measuring their equivariance
  and equivalence.
\newblock In \emph{CVPR}, 2015.

\bibitem[Lester et~al.(2021)Lester, Al-Rfou, and
  Constant]{lester2021prompttuning}
Brian Lester, Rami Al-Rfou, and Noah Constant.
\newblock The power of scale for parameter-efficient prompt tuning.
\newblock \emph{arXiv preprint arXiv:2104.08691}, 2021.

\bibitem[Li et~al.(2025{\natexlab{a}})Li, Sun, Liu, Wang, Wu, Yu, Chen,
  Barsoum, Chen, and Liu]{li2025latent}
Bangzheng Li, Ximeng Sun, Jiang Liu, Ze~Wang, Jialian Wu, Xiaodong Yu, Hao
  Chen, Emad Barsoum, Muhao Chen, and Zicheng Liu.
\newblock Latent visual reasoning.
\newblock \emph{arXiv preprint arXiv:2509.24251}, 2025{\natexlab{a}}.

\bibitem[Li \& Shi(2026)Li and Shi]{li2026latentbridgecontinuousslowfast}
Bojie Li and Noah Shi.
\newblock The latent bridge: A continuous slow-fast channel for real-time game
  agents, 2026.
\newblock URL \url{https://arxiv.org/abs/2606.24470}.

\bibitem[Li et~al.(2025{\natexlab{b}})Li, Zhang, and Zong]{li2025tokalign}
Chong Li, Jiajun Zhang, and Chengqing Zong.
\newblock Tokalign: Efficient vocabulary adaptation via token alignment.
\newblock \emph{arXiv preprint arXiv:2506.03523}, 2025{\natexlab{b}}.
\newblock URL \url{https://arxiv.org/abs/2506.03523}.

\bibitem[Li et~al.(2023{\natexlab{a}})Li, Hammoud, Itani, Khizbullin, and
  Ghanem]{li2023camel}
Guohao Li, Hasan Hammoud, Hani Itani, Dmitrii Khizbullin, and Bernard Ghanem.
\newblock Camel: Communicative agents for "mind" exploration of large language
  model society.
\newblock In \emph{Advances in Neural Information Processing Systems 36: Annual
  Conference on Neural Information Processing Systems 2023 (NeurIPS 2023)},
  2023{\natexlab{a}}.
\newblock URL
  \url{https://proceedings.neurips.cc/paper/2023/hash/a3621ee907def47c1b952ade25c67698-Abstract-Conference.html}.

\bibitem[Li et~al.(2023{\natexlab{b}})Li, Li, Savarese, and Hoi]{li2023blip2}
Junnan Li, Dongxu Li, Silvio Savarese, and Steven Hoi.
\newblock Blip-2: Bootstrapping language-image pre-training with frozen image
  encoders and large language models, 2023{\natexlab{b}}.

\bibitem[Li \& Liang(2021)Li and Liang]{li2021prefixtuning}
Xiang~Lisa Li and Percy Liang.
\newblock Prefix-tuning: Optimizing continuous prompts for generation.
\newblock \emph{arXiv preprint arXiv:2101.00190}, 2021.

\bibitem[Li et~al.(2026)Li, An, and Du]{li2026latentleadsbetterrelay}
Yiping Li, Zhiyu An, and Wan Du.
\newblock When less latent leads to better relay: Information-preserving
  compression for latent multi-agent llm collaboration, 2026.
\newblock URL \url{https://arxiv.org/abs/2604.13349}.

\bibitem[Li et~al.(2025{\natexlab{c}})Li, Wu, Guo, Sun, and
  Han]{li2025embodied}
Zhuo Li, Weiran Wu, Yunlong Guo, Jian Sun, and Qing-Long Han.
\newblock Embodied multi-agent systems: A review.
\newblock \emph{IEEE/CAA Journal of Automatica Sinica}, 12\penalty0
  (6):\penalty0 1095--1116, 2025{\natexlab{c}}.

\bibitem[Li et~al.(2025{\natexlab{d}})Li, Zhang, Han, Liu, Xie, Zhang, Choi,
  Zou, and Lu]{li2025flow}
Zhuofeng Li, Haoxiang Zhang, Seungju Han, Sheng Liu, Jianwen Xie, Yu~Zhang,
  Yejin Choi, James Zou, and Pan Lu.
\newblock In-the-flow agentic system optimization for effective planning and
  tool use.
\newblock \emph{arXiv preprint arXiv:2510.05592}, 2025{\natexlab{d}}.

\bibitem[Liu et~al.(2023{\natexlab{a}})Liu, Jiang, Zhang, Liu, Zhang, Biswas,
  and Stone]{liu2023llmpempoweringlargelanguage}
Bo~Liu, Yuqian Jiang, Xiaohan Zhang, Qiang Liu, Shiqi Zhang, Joydeep Biswas,
  and Peter Stone.
\newblock Llm+p: Empowering large language models with optimal planning
  proficiency, 2023{\natexlab{a}}.
\newblock URL \url{https://arxiv.org/abs/2304.11477}.

\bibitem[Liu et~al.(2023{\natexlab{b}})Liu, Xia, Wang, and Zhang]{codeplus}
Jiawei Liu, Chunqiu~Steven Xia, Yuyao Wang, and Lingming Zhang.
\newblock Is your code generated by chatgpt really correct? rigorous evaluation
  of large language models for code generation.
\newblock \emph{Advances in Neural Information Processing Systems},
  36:\penalty0 21558--21572, 2023{\natexlab{b}}.

\bibitem[Liu et~al.(2024)Liu, Pfeiffer, Wu, Xie, and
  Szlam]{liu2024deliberation}
Luyang Liu, Jonas Pfeiffer, Jiaxing Wu, Jun Xie, and Arthur Szlam.
\newblock Deliberation in latent space via differentiable cache augmentation.
\newblock \emph{arXiv preprint arXiv:2412.17747}, 2024.

\bibitem[Liu et~al.(2026)Liu, Yu, Fredrikson, Wang, and
  Gao]{liu2026trojanvocabularystealthysabotage}
Xiaoze Liu, Weichen Yu, Matt Fredrikson, Xiaoqian Wang, and Jing Gao.
\newblock When the same coefficients reach different places: Asymmetric
  realizability in transplanting tokenizers across large language models, 2026.
\newblock URL \url{https://arxiv.org/abs/2601.00065}.

\bibitem[Liu(2026)]{liu2026tokensunifiedframeworklatent}
Yingzhuo Liu.
\newblock Beyond tokens: a unified framework for latent communication in
  llm-based multi-agent systems, 2026.
\newblock URL \url{https://arxiv.org/abs/2606.05711}.

\bibitem[Lu \& Deshpande(2026{\natexlab{a}})Lu and
  Deshpande]{lu2026alignmentrole}
Wenhao Lu and Spursh Deshpande.
\newblock On the role of learned alignment matrices in {LatentMAS}.
\newblock In \emph{Second Workshop on Compositional Learning: Safety,
  Interpretability, and Agents at ICML}, 2026{\natexlab{a}}.
\newblock URL \url{https://openreview.net/forum?id=manXhfpRH3}.

\bibitem[Lu \& Deshpande(2026{\natexlab{b}})Lu and
  Deshpande]{lu2026latentrepresent}
Wenhao Lu and Spursh Deshpande.
\newblock What do latent agents actually represent? interpreting hidden-state
  communication in multi-agent systems.
\newblock In \emph{Second Workshop on Compositional Learning: Safety,
  Interpretability, and Agents at ICML}, 2026{\natexlab{b}}.
\newblock URL \url{https://openreview.net/forum?id=UJdeo8SM7F}.

\bibitem[Marafioti et~al.(2025)Marafioti, Zohar, Farr{\'e}, Noyan, Bakouch,
  Cuenca, Zakka, Allal, Lozhkov, Tazi, Srivastav, Lochner, Larcher, Morlon,
  Tunstall, von Werra, and Wolf]{marafioti2025smolvlmredefiningsmallefficient}
Andr{\'e}s Marafioti, Orr Zohar, Miquel Farr{\'e}, Merve Noyan, Elie Bakouch,
  Pedro Cuenca, Cyril Zakka, Loubna~Ben Allal, Anton Lozhkov, Nouamane Tazi,
  Vaibhav Srivastav, Joshua Lochner, Hugo Larcher, Mathieu Morlon, Lewis
  Tunstall, Leandro von Werra, and Thomas Wolf.
\newblock Smolvlm: Redefining small and efficient multimodal models, 2025.
\newblock URL \url{https://arxiv.org/abs/2504.05299}.

\bibitem[Maxwell-Jia(2024)]{aime24}
Maxwell-Jia.
\newblock {AIME 2024} dataset.
\newblock \url{https://huggingface.co/datasets/Maxwell-Jia/AIME_2024}, 2024.

\bibitem[Minixhofer et~al.(2024)Minixhofer, Ponti, and
  Vuli{\'c}]{minixhofer2024zero}
Benjamin Minixhofer, Edoardo~Maria Ponti, and Ivan Vuli{\'c}.
\newblock Zero-shot tokenizer transfer.
\newblock In \emph{Advances in Neural Information Processing Systems},
  volume~37, 2024.

\bibitem[Minixhofer et~al.(2025)Minixhofer, Vuli{\'c}, and
  Ponti]{minixhofer2025universal}
Benjamin Minixhofer, Ivan Vuli{\'c}, and Edoardo~Maria Ponti.
\newblock Universal cross-tokenizer distillation via approximate likelihood
  matching.
\newblock \emph{arXiv preprint arXiv:2503.20083}, 2025.

\bibitem[Moroni et~al.(2025)Moroni, Puccetti, Huguet~Cabot, Bejgu, Miaschi,
  Barba, Dell'Orletta, Esuli, and Navigli]{moroni2025optimizing}
Luca Moroni, Giovanni Puccetti, Pere-Llu\'is Huguet~Cabot, Andrei~Stefan Bejgu,
  Alessio Miaschi, Edoardo Barba, Felice Dell'Orletta, Andrea Esuli, and
  Roberto Navigli.
\newblock Optimizing llms for italian: Reducing token fertility and enhancing
  efficiency through vocabulary adaptation.
\newblock In \emph{Findings of the Association for Computational Linguistics:
  NAACL 2025}, pp.\  6661--6675, Albuquerque, New Mexico, April 2025.
  Association for Computational Linguistics.
\newblock \doi{10.18653/v1/2025.findings-naacl.371}.
\newblock URL \url{https://aclanthology.org/2025.findings-naacl.371/}.

\bibitem[Moschella(2024)]{moschella2024latentcommunicationartificialneural}
Luca Moschella.
\newblock Latent communication in artificial neural networks, 2024.
\newblock URL \url{https://arxiv.org/abs/2406.11014}.

\bibitem[Mundra et~al.(2024)Mundra, Khandavally, Dabre, Puduppully,
  Kunchukuttan, and Khapra]{mundra2024empirical}
Nandini Mundra, Aditya Nanda~Kishore Khandavally, Raj Dabre, Ratish Puduppully,
  Anoop Kunchukuttan, and Mitesh~M. Khapra.
\newblock An empirical comparison of vocabulary expansion and initialization
  approaches for language models.
\newblock In \emph{Proceedings of the 28th Conference on Computational Natural
  Language Learning (CoNLL)}, pp.\  84--104, Miami, FL, USA, nov 2024.
  Association for Computational Linguistics.
\newblock \doi{10.18653/v1/2024.conll-1.8}.
\newblock URL \url{https://aclanthology.org/2024.conll-1.8/}.

\bibitem[{nhminle}(2026)]{latentmashybrid2026}
{nhminle}.
\newblock Latentmas-hybrid.
\newblock \url{https://github.com/nhminle/LatentMAS-Hybrid}, 2026.
\newblock Public fork of LatentMAS. Cross-model realignment implementation at
  commit \texttt{6db9551}. Accessed September 10, 2026.

\bibitem[Packer et~al.(2023)Packer, Wooders, Lin, Fang, Patil, Stoica, and
  Gonzalez]{packer2023memgpt}
Charles Packer, Sarah Wooders, Kevin Lin, Vivian Fang, Shishir~G Patil, Ion
  Stoica, and Joseph~E Gonzalez.
\newblock Memgpt: Towards llms as operating systems.
\newblock \emph{arXiv preprint arXiv:2310.08560}, 2023.

\bibitem[Pan et~al.(2025)Pan, Wu, Jiang, Luo, Cheng, Li, Yang, Lin, Zhao, Qiu,
  and Gao]{pan2025secom}
Zhuoshi Pan, Qianhui Wu, Huiqiang Jiang, Xufang Luo, Hao Cheng, Dongsheng Li,
  Yuqing Yang, Chin-Yew Lin, H.~Vicky Zhao, Lili Qiu, and Jianfeng Gao.
\newblock Secom: On memory construction and retrieval for personalized
  conversational agents.
\newblock In \emph{The Thirteenth International Conference on Learning
  Representations}, 2025.
\newblock URL \url{https://openreview.net/forum?id=xKDZAW0He3}.

\bibitem[Pappu et~al.(2026)Pappu, El, Cao, di~Nolfo, Sun, Cao, and
  Zou]{pappu2026multiagentteamsholdexperts}
Aneesh Pappu, Batu El, Hancheng Cao, Carmelo di~Nolfo, Yanchao Sun, Meng Cao,
  and James Zou.
\newblock Multi-agent teams hold experts back, 2026.
\newblock URL \url{https://arxiv.org/abs/2602.01011}.

\bibitem[Park et~al.(2023{\natexlab{a}})Park, O'Brien, Cai, Morris, Liang, and
  Bernstein]{park2023generative}
Joon~Sung Park, Joseph O'Brien, Carrie~Jun Cai, Meredith~Ringel Morris, Percy
  Liang, and Michael~S Bernstein.
\newblock Generative agents: Interactive simulacra of human behavior.
\newblock In \emph{Proceedings of the 36th annual acm symposium on user
  interface software and technology}, pp.\  1--22, 2023{\natexlab{a}}.

\bibitem[Park et~al.(2023{\natexlab{b}})Park, Choe, and Veitch]{park2023linear}
Kiho Park, Yo~Joong Choe, and Victor Veitch.
\newblock The linear representation hypothesis and the geometry of large
  language models.
\newblock \emph{arXiv preprint arXiv:2311.03658}, 2023{\natexlab{b}}.

\bibitem[Pezeshkpour et~al.(2024)Pezeshkpour, Kandogan, Bhutani, Rahman,
  Mitchell, and Hruschka]{pezeshkpour2024reasoning}
Pouya Pezeshkpour, Eser Kandogan, Nikita Bhutani, Sajjadur Rahman, Tom
  Mitchell, and Estevam Hruschka.
\newblock Reasoning capacity in multi-agent systems: Limitations, challenges
  and human-centered solutions.
\newblock \emph{arXiv preprint arXiv:2402.01108}, 2024.

\bibitem[Pfeiffer et~al.(2021)Pfeiffer, Vuli{\'c}, Gurevych, and
  Ruder]{pfeiffer2021unks}
Jonas Pfeiffer, Ivan Vuli{\'c}, Iryna Gurevych, and Sebastian Ruder.
\newblock Unks everywhere: Adapting multilingual language models to new
  scripts.
\newblock In \emph{Proceedings of the 2021 Conference on Empirical Methods in
  Natural Language Processing (EMNLP)}, pp.\  10186--10203, 2021.

\bibitem[Pham et~al.(2024)Pham, Liu, Yang, Chen, Liu, Yuan, Plummer, Wang, and
  Yang]{pham2024cipher}
Chau Pham, Boyi Liu, Yingxiang Yang, Zhengyu Chen, Tianyi Liu, Jianbo Yuan,
  Bryan~A. Plummer, Zhaoran Wang, and Hongxia Yang.
\newblock Let models speak ciphers: Multiagent debate through embeddings.
\newblock In \emph{The Twelfth International Conference on Learning
  Representations (ICLR)}, 2024.
\newblock URL \url{https://openreview.net/forum?id=sehRvaIPQQ}.

\bibitem[Qu et~al.(2025)Qu, Li, Su, Sun, Yan, Liu, Cui, Liu, Liang, He, Li,
  Wei, Shao, Lu, Zhang, Hua, Zhou, and Cheng]{qu2025survey}
Xiaoye Qu, Yafu Li, Zhaochen Su, Weigao Sun, Jianhao Yan, Dongrui Liu, Ganqu
  Cui, Daizong Liu, Shuxian Liang, Junxian He, Peng Li, Wei Wei, Jing Shao,
  Chaochao Lu, Yue Zhang, Xian-Sheng Hua, Bowen Zhou, and Yu~Cheng.
\newblock A survey of efficient reasoning for large reasoning models: Language,
  multimodality, and beyond.
\newblock \emph{arXiv preprint arXiv:2503.21614}, 2025.

\bibitem[Radford et~al.(2021)Radford, Kim, Hallacy, Ramesh, Goh, Agarwal,
  Sastry, Askell, Mishkin, Clark, Krueger, and Sutskever]{radford2021clip}
Alec Radford, Jong~Wook Kim, Chris Hallacy, Aditya Ramesh, Gabriel Goh,
  Sandhini Agarwal, Girish Sastry, Amanda Askell, Pamela Mishkin, Jack Clark,
  Gretchen Krueger, and Ilya Sutskever.
\newblock Learning transferable visual models from natural language
  supervision.
\newblock In \emph{Proceedings of the 38th International Conference on Machine
  Learning}, volume 139 of \emph{Proceedings of Machine Learning Research},
  pp.\  8748--8763. PMLR, 2021.

\bibitem[Ramesh \& Li(2025)Ramesh and Li]{ramesh2025activations}
Vignav Ramesh and Kenneth Li.
\newblock Communicating activations between language model agents.
\newblock In \emph{Proceedings of the 42nd International Conference on Machine
  Learning (ICML)}, 2025.
\newblock URL \url{https://proceedings.mlr.press/v267/ramesh25a.html}.

\bibitem[Rein et~al.(2023)Rein, Hou, Stickland, Petty, Pang, Dirani, Michael,
  and Bowman]{gpqa}
David Rein, Betty~Li Hou, Asa~Cooper Stickland, Jackson Petty, Richard~Yuanzhe
  Pang, Julien Dirani, Julian Michael, and Samuel~R. Bowman.
\newblock Gpqa: A graduate-level google-proof q\&a benchmark, 2023.
\newblock URL \url{https://arxiv.org/abs/2311.12022}.

\bibitem[Remy et~al.(2024)Remy, Delobelle, Avetisyan, Khabibullina, de~Lhoneux,
  and Demeester]{remy2024trans}
Fran\c{c}ois Remy, Pieter Delobelle, Hayastan Avetisyan, Alfiya Khabibullina,
  Miryam de~Lhoneux, and Thomas Demeester.
\newblock Trans-tokenization and cross-lingual vocabulary transfers: Language
  adaptation of llms for low-resource nlp.
\newblock \emph{arXiv preprint arXiv:2408.04303}, 2024.
\newblock URL \url{https://arxiv.org/abs/2408.04303}.
\newblock Accepted at COLM 2024.

\bibitem[Remy et~al.(2023)Remy, Delobelle, Berendt, Demuynck, and
  Demeester]{remy2023tik}
Fran{\c{c}}ois Remy, Pieter Delobelle, Bettina Berendt, Kris Demuynck, and
  Thomas Demeester.
\newblock Tik-to-tok: Translating language models one token at a time: An
  embedding initialization strategy for efficient language adaptation.
\newblock \emph{arXiv preprint arXiv:2310.03477}, 2023.

\bibitem[Rossi et~al.(2026)Rossi, Raghunath, and
  Wu]{rossi2026latentcacheflowmodeltomodel}
Maximillian Rossi, Prajwal Raghunath, and Eugene Wu.
\newblock Latent cache flow: Model-to-model communication without text, 2026.
\newblock URL \url{https://arxiv.org/abs/2605.22863}.

\bibitem[Ruan et~al.(2023)Ruan, Chen, Zhang, Xu, Bao, Du, Shi, Mao, Zeng, and
  Zhao]{ruan2023tptu}
Jingqing Ruan, Yihong Chen, Bin Zhang, Zhiwei Xu, Tianpeng Bao, Guoqing Du,
  Shiwei Shi, Hangyu Mao, Xingyu Zeng, and Rui Zhao.
\newblock Tptu: large language model-based ai agents for task planning and tool
  usage.
\newblock \emph{arXiv preprint arXiv:2308.03427}, 2023.

\bibitem[Rust et~al.(2021)Rust, Pfeiffer, Vuli{\'c}, Ruder, and
  Gurevych]{rust2021good}
Phillip Rust, Jonas Pfeiffer, Ivan Vuli{\'c}, Sebastian Ruder, and Iryna
  Gurevych.
\newblock How good is your tokenizer? on the monolingual performance of
  multilingual language models.
\newblock In \emph{Proceedings of the 59th Annual Meeting of the Association
  for Computational Linguistics}, 2021.

\bibitem[Sarker et~al.(2026)Sarker, Halpern, Staylor, Alsaadi, von Laszewski,
  Cheng, Jha, and Fox]{sarker2026aaflowstatefuloperatorabstraction}
Arup~Kumar Sarker, Alexander~James Halpern, Mills Staylor, Aymen Alsaadi,
  Gregor von Laszewski, Yue Cheng, Shantenu Jha, and Geoffrey Fox.
\newblock [aaflow+] stateful operator abstraction with zero-copy distributed kv
  cache orchestration for multi-agent workflows, 2026.
\newblock URL \url{https://arxiv.org/abs/2607.10987}.

\bibitem[Sharthak et~al.(2025)Sharthak, Pahalwan, Kamath, and
  Shirawalmath]{sharthak2025achieving}
Shaurya Sharthak, Vinayak Pahalwan, Adithya Kamath, and Adarsh Shirawalmath.
\newblock Achieving tokenizer flexibility in language models through heuristic
  adaptation and supertoken learning.
\newblock \emph{arXiv preprint arXiv:2505.09738}, 2025.
\newblock URL \url{https://arxiv.org/abs/2505.09738}.

\bibitem[Shi et~al.(2026{\natexlab{a}})Shi, Chiesa, Jr., and
  Kostic]{shi2026kvcommenablingefficientllm}
Xiangyu Shi, Marco Chiesa, Gerald Q.~Maguire Jr., and Dejan Kostic.
\newblock Kvcomm: Enabling efficient llm communication through selective kv
  sharing, 2026{\natexlab{a}}.
\newblock URL \url{https://arxiv.org/abs/2510.03346}.

\bibitem[Shi et~al.(2026{\natexlab{b}})Shi, Xie, Sun, Chen, Zhang, Yun, Wan,
  Zhang, Lo, and Gu]{shi2026codeocreffectivenessvisionlanguage}
Yuling Shi, Chaoxiang Xie, Zhensu Sun, Yeheng Chen, Chenxu Zhang, Longfei Yun,
  Chengcheng Wan, Hongyu Zhang, David Lo, and Xiaodong Gu.
\newblock Codeocr: On the effectiveness of vision language models in code
  understanding, 2026{\natexlab{b}}.
\newblock URL \url{https://arxiv.org/abs/2602.01785}.

\bibitem[Shkolnikov(2026)]{shkolnikov2026agentmemorypromptpersistent}
Yakov~Pyotr Shkolnikov.
\newblock Agent memory below the prompt: Persistent q4 kv cache for multi-agent
  llm inference on edge devices, 2026.
\newblock URL \url{https://arxiv.org/abs/2603.04428}.

\bibitem[Sui et~al.(2025)Sui, Chuang, Wang, Zhang, Zhang, Yuan, Liu, Wen,
  Zhong, Zou, Chen, and Hu]{sui2025stop}
Yang Sui, Yu-Neng Chuang, Guanchu Wang, Jiamu Zhang, Tianyi Zhang, Jiayi Yuan,
  Hongyi Liu, Andrew Wen, Shaochen Zhong, Na~Zou, Hanjie Chen, and Xia Hu.
\newblock Stop overthinking: A survey on efficient reasoning for large language
  models.
\newblock \emph{arXiv preprint arXiv:2503.16419}, 2025.

\bibitem[Tai et~al.(2020)Tai, Kung, Dong, Comiter, and Kuo]{tai2020exbert}
Wen Tai, H.~T. Kung, Xin Dong, Marcus Comiter, and Chang-Fu Kuo.
\newblock exbert: Extending pre-trained models with domain-specific vocabulary
  under constrained training resources.
\newblock In Trevor Cohn, Yulan He, and Yang Liu (eds.), \emph{Findings of the
  Association for Computational Linguistics: EMNLP 2020}, pp.\  1433--1439,
  Online, nov 2020. Association for Computational Linguistics.
\newblock \doi{10.18653/v1/2020.findings-emnlp.129}.
\newblock URL \url{https://aclanthology.org/2020.findings-emnlp.129/}.

\bibitem[Tang et~al.(2025)Tang, Su, Zhou, Liu, Zhang, Ma, and
  Ai]{tang2025augmentingmultiagentcommunicationstate}
Yichen Tang, Weihang Su, Yujia Zhou, Yiqun Liu, Min Zhang, Shaoping Ma, and
  Qingyao Ai.
\newblock Augmenting multi-agent communication with state delta trajectory,
  2025.
\newblock URL \url{https://arxiv.org/abs/2506.19209}.

\bibitem[Tao et~al.(2024)Tao, Zhou, Wang, Zhang, Zhang, and
  Cheng]{tao2024magis}
Wei Tao, Yucheng Zhou, Yanlin Wang, Wenqiang Zhang, Hongyu Zhang, and Yu~Cheng.
\newblock Magis: Llm-based multi-agent framework for github issue resolution.
\newblock \emph{Advances in Neural Information Processing Systems},
  37:\penalty0 51963--51993, 2024.

\bibitem[Team(2025)]{gemmateam2025gemma3technicalreport}
Gemma Team.
\newblock Gemma 3 technical report, 2025.
\newblock URL \url{https://arxiv.org/abs/2503.19786}.

\bibitem[Tran et~al.(2025)Tran, Dao, Nguyen, Pham, O'Sullivan, and
  Nguyen]{tran2025multi}
Khanh-Tung Tran, Dung Dao, Minh-Duong Nguyen, Quoc-Viet Pham, Barry O'Sullivan,
  and Hoang~D Nguyen.
\newblock Multi-agent collaboration mechanisms: A survey of llms.
\newblock \emph{arXiv preprint arXiv:2501.06322}, 2025.

\bibitem[Vernikos \& Popescu-Belis(2021)Vernikos and
  Popescu-Belis]{vernikos2021subword}
Giorgos Vernikos and Andrei Popescu-Belis.
\newblock Subword mapping and anchoring across languages.
\newblock \emph{arXiv preprint arXiv:2109.04556}, 2021.

\bibitem[Wang et~al.(2026{\natexlab{a}})Wang, Huang, Sun, Wei, and
  Wu]{wang2026sightmindunveilinglatent}
Chenxi Wang, Ruiyang Huang, Jiayan Sun, Lei Wei, and Yifan Wu.
\newblock Out of sight, not out of mind: Unveiling latent attack in
  latent-based multi-agent systems, 2026{\natexlab{a}}.
\newblock URL \url{https://arxiv.org/abs/2605.28214}.

\bibitem[Wang et~al.(2024{\natexlab{a}})Wang, Wang, Athiwaratkun, Zhang, and
  Zou]{wang2024mixtureofagentsenhanceslargelanguage}
Junlin Wang, Jue Wang, Ben Athiwaratkun, Ce~Zhang, and James Zou.
\newblock Mixture-of-agents enhances large language model capabilities,
  2024{\natexlab{a}}.
\newblock URL \url{https://arxiv.org/abs/2406.04692}.

\bibitem[Wang et~al.(2024{\natexlab{b}})Wang, Ma, Feng, Zhang, Yang, Zhang,
  Chen, Tang, Chen, Lin, Zhao, Wei, and Wen]{wang2024survey}
Lei Wang, Chen Ma, Xueyang Feng, Zeyu Zhang, Hao Yang, Jingsen Zhang, Zhiyuan
  Chen, Jiakai Tang, Xu~Chen, Yankai Lin, Wayne~Xin Zhao, Zhewei Wei, and
  Jirong Wen.
\newblock A survey on large language model based autonomous agents.
\newblock \emph{Frontiers of Computer Science}, 18\penalty0 (6):\penalty0
  186345, 2024{\natexlab{b}}.

\bibitem[Wang et~al.(2025{\natexlab{a}})Wang, Shi, Wang, Zhang, Wan, Gai, Ying,
  and Wang]{wang2025monet}
Qixun Wang, Yang Shi, Yifei Wang, Yuanxing Zhang, Pengfei Wan, Kun Gai,
  Xianghua Ying, and Yisen Wang.
\newblock Monet: Reasoning in latent visual space beyond images and language.
\newblock \emph{arXiv preprint arXiv:2511.21395}, 2025{\natexlab{a}}.

\bibitem[Wang et~al.(2025{\natexlab{b}})Wang, Wang, Xue, Pang, Liu, Chen, Qiu,
  Wong, Ji, and Wong]{wang2025harnessing}
Rui Wang, Hongru Wang, Boyang Xue, Jianhui Pang, Shudong Liu, Yi~Chen, Jiahao
  Qiu, Derek~Fai Wong, Heng Ji, and Kam-Fai Wong.
\newblock Harnessing the reasoning economy: A survey of efficient reasoning for
  large language models.
\newblock \emph{arXiv preprint arXiv:2503.24377}, 2025{\natexlab{b}}.

\bibitem[Wang et~al.(2026{\natexlab{b}})Wang, Li, Li, Yang, Tang, and
  Wei]{wang2026renderofthoughtrenderingtextualchainofthought}
Yifan Wang, Shiyu Li, Peiming Li, Xiaochen Yang, Yang Tang, and Zheng Wei.
\newblock Render-of-thought: Rendering textual chain-of-thought as images for
  visual latent reasoning, 2026{\natexlab{b}}.
\newblock URL \url{https://arxiv.org/abs/2601.14750}.

\bibitem[Wang et~al.(2024{\natexlab{c}})Wang, Gao, Chen, Jiang, Li, Yang, Yin,
  Li, Li, Yin, Shang, and McAuley]{wang2024memoryllm}
Yu~Wang, Yifan Gao, Xiusi Chen, Haoming Jiang, Shiyang Li, Jingfeng Yang,
  Qingyu Yin, Zheng Li, Xian Li, Bing Yin, Jingbo Shang, and Julian McAuley.
\newblock {MEMORYLLM:} towards self-updatable large language models.
\newblock In \emph{Proceedings of the 41st International Conference on Machine
  Learning}, volume 235 of \emph{Proceedings of Machine Learning Research},
  pp.\  50453--50466. PMLR, 2024{\natexlab{c}}.

\bibitem[Wang et~al.(2025{\natexlab{c}})Wang, Moriyama, Wang, Gangopadhyay, and
  Takamatsu]{wang2025talk}
Zhao Wang, Sota Moriyama, Wei-Yao Wang, Briti Gangopadhyay, and Shingo
  Takamatsu.
\newblock Talk structurally, act hierarchically: A collaborative framework for
  llm multi-agent systems.
\newblock \emph{arXiv preprint arXiv:2502.11098}, 2025{\natexlab{c}}.

\bibitem[Wang et~al.(2026{\natexlab{c}})Wang, Zhao, and
  Ding]{wang2026medlatentdxlatentmultiagentcommunication}
Ziqing Wang, Lili Zhao, and Kaize Ding.
\newblock Medlatentdx: Latent multi-agent communication for cross-hospital
  rare-disease diagnosis, 2026{\natexlab{c}}.
\newblock URL \url{https://arxiv.org/abs/2606.13945}.

\bibitem[Wei et~al.(2025)Wei, Sun, and
  Li]{wei2025deepseekocrcontextsopticalcompression}
Haoran Wei, Yaofeng Sun, and Yukun Li.
\newblock Deepseek-ocr: Contexts optical compression, 2025.
\newblock URL \url{https://arxiv.org/abs/2510.18234}.

\bibitem[Wenzel(2026)]{wenzel2026latentcommunicationlanguagemodel}
Markus Wenzel.
\newblock Latent communication between language model agents: Channels,
  alignment, and the limits of text, 2026.
\newblock URL \url{https://arxiv.org/abs/2607.14103}.

\bibitem[Wortsman et~al.(2022)Wortsman, Ilharco, Gadre, Roelofs, Gontijo-Lopes,
  Morcos, Namkoong, Farhadi, Carmon, Kornblith, and Schmidt]{wortsman2022model}
Mitchell Wortsman, Gabriel Ilharco, Samir~Yitzhak Gadre, Rebecca Roelofs,
  Raphael Gontijo-Lopes, Ari~S. Morcos, Hongseok Namkoong, Ali Farhadi, Yair
  Carmon, Simon Kornblith, and Ludwig Schmidt.
\newblock Model soups: averaging weights of multiple fine-tuned models improves
  accuracy without increasing inference time.
\newblock In \emph{International conference on machine learning}, pp.\
  23965--23998. PMLR, 2022.

\bibitem[Wu \& Zhu(2026)Wu and Zhu]{wu2026posthocsparsecodinglatent}
Di~Wu and Xiaohui Zhu.
\newblock Post-hoc sparse coding of latent communication between
  vision-language model agents, 2026.
\newblock URL \url{https://arxiv.org/abs/2608.10198}.

\bibitem[Wu et~al.(2025)Wu, Li, Wei, Li, Ding, and Gao]{wu2025talk}
Feijie Wu, Zitao Li, Fei Wei, Yaliang Li, Bolin Ding, and Jing Gao.
\newblock Talk to right specialists: Routing and planning in multi-agent system
  for question answering.
\newblock \emph{arXiv preprint arXiv:2501.07813}, 2025.

\bibitem[Wu et~al.(2024)Wu, Bansal, Zhang, Wu, Li, Zhu, Jiang, Zhang, Zhang,
  Liu, Awadallah, White, Burger, and Wang]{wu2024autogen}
Qingyun Wu, Gagan Bansal, Jieyu Zhang, Yiran Wu, Beibin Li, Erkang Zhu,
  Li~Jiang, Xiaoyun Zhang, Shaokun Zhang, Jiale Liu, Ahmed~Hassan Awadallah,
  Ryen~W White, Doug Burger, and Chi Wang.
\newblock Autogen: Enabling next-gen {LLM} applications via multi-agent
  conversation.
\newblock In \emph{First Conference on Language Modeling}, 2024.

\bibitem[Yamaguchi et~al.(2024)Yamaguchi, Villavicencio, and
  Aletras]{yamaguchi2024empirical}
Atsuki Yamaguchi, Aline Villavicencio, and Nikolaos Aletras.
\newblock An empirical study on cross-lingual vocabulary adaptation for
  efficient language model inference.
\newblock \emph{arXiv preprint arXiv:2402.10712}, 2024.

\bibitem[Yamaguchi et~al.(2026)Yamaguchi, Villavicencio, and
  Aletras]{yamaguchi2025how}
Atsuki Yamaguchi, Aline Villavicencio, and Nikolaos Aletras.
\newblock How can we effectively expand the vocabulary of {LLMs} with 0.01{GB}
  of target language text?
\newblock \emph{Computational Linguistics}, 52\penalty0 (1):\penalty0 295--330,
  2026.

\bibitem[Yan et~al.(2025)Yan, Zhou, Zhang, Zhang, Zhou, Miao, Li, Li, and
  Zhang]{yan2025beyond}
Bingyu Yan, Zhibo Zhou, Litian Zhang, Lian Zhang, Ziyi Zhou, Dezhuang Miao,
  Zhoujun Li, Chaozhuo Li, and Xiaoming Zhang.
\newblock Beyond self-talk: A communication-centric survey of llm-based
  multi-agent systems.
\newblock \emph{arXiv preprint arXiv:2502.14321}, 2025.

\bibitem[Yang \& Eshraghian(2025)Yang and
  Eshraghian]{yang2025directsemanticcommunicationlarge}
Fu-Chun Yang and Jason Eshraghian.
\newblock Direct semantic communication between large language models via
  vector translation, 2025.
\newblock URL \url{https://arxiv.org/abs/2511.03945}.

\bibitem[Yang et~al.(2025)Yang, Chen, Guo, Chen, Lin, Hu, Hu, Wu, and
  Wang]{medqa}
Hang Yang, Hao Chen, Hui Guo, Yineng Chen, Ching-Sheng Lin, Shu Hu, Jinrong Hu,
  Xi~Wu, and Xin Wang.
\newblock Llm-medqa: Enhancing medical question answering through case studies
  in large language models.
\newblock \emph{arXiv preprint arXiv:2501.05464}, 2025.
\newblock Submitted 31 Dec 2024 (v1); revised 18 Jan 2025 (v2).

\bibitem[Yang et~al.(2026)Yang, Huang, Zhang, Wang, Yu, and
  Lee]{yang2026xbridgeentitygroundedlatentbridge}
Wooseong Yang, Wei-Chieh Huang, Weizhi Zhang, Yu~Wang, Philip~S. Yu, and
  Junhyun Lee.
\newblock Xbridge: Entity-grounded latent bridge for heterogeneous llm
  communication, 2026.
\newblock URL \url{https://arxiv.org/abs/2608.11676}.

\bibitem[Yang et~al.(2024)Yang, Peng, Wang, Wen, and Zhang]{yang2024llm}
Yingxuan Yang, Qiuying Peng, Jun Wang, Ying Wen, and Weinan Zhang.
\newblock Llm-based multi-agent systems: Techniques and business perspectives.
\newblock \emph{arXiv preprint arXiv:2411.14033}, 2024.

\bibitem[Yao et~al.(2025)Yao, Zhang, Huang, Zhang, Wang, Fang, Zhu, Jing, Liu,
  Li, and Tao]{yao2025survey}
Huanjin Yao, Ruifei Zhang, Jiaxing Huang, Jingyi Zhang, Yibo Wang, Bo~Fang,
  Ruolin Zhu, Yongcheng Jing, Shunyu Liu, Guanbin Li, and Dacheng Tao.
\newblock A survey on agentic multimodal large language models.
\newblock \emph{arXiv preprint arXiv:2510.10991}, 2025.

\bibitem[Ye et~al.(2025{\natexlab{a}})Ye, Gao, Ma, Wang, Fu, Chung, Lin, Liu,
  Zhang, Zhuo, and Chen]{ye2025kvcomm}
Hancheng Ye, Zhengqi Gao, Mingyuan Ma, Qinsi Wang, Yuzhe Fu, Ming-Yu Chung,
  Yueqian Lin, Zhijian Liu, Jianyi Zhang, Danyang Zhuo, and Yiran Chen.
\newblock Kvcomm: Online cross-context kv-cache communication for efficient
  llm-based multi-agent systems.
\newblock \emph{arXiv preprint arXiv:2510.12872}, 2025{\natexlab{a}}.

\bibitem[Ye et~al.(2025{\natexlab{b}})Ye, Liu, Wu, Pang, Yin, Bai, and
  Chen]{ye2025xmasbuildingmultiagentsystems}
Rui Ye, Xiangrui Liu, Qimin Wu, Xianghe Pang, Zhenfei Yin, Lei Bai, and Siheng
  Chen.
\newblock X-mas: Towards building multi-agent systems with heterogeneous llms,
  2025{\natexlab{b}}.
\newblock URL \url{https://arxiv.org/abs/2505.16997}.

\bibitem[Yi et~al.(2026)Yi, Mueller, and
  Lee]{yi2026latentagentsposttrainingprocedure}
John Seon~Keun Yi, Aaron Mueller, and Dokyun Lee.
\newblock Latent agents: A post-training procedure for internalized multi-agent
  debate, 2026.
\newblock URL \url{https://arxiv.org/abs/2604.24881}.

\bibitem[Yu et~al.(2026{\natexlab{a}})Yu, Xu, Chen, Yin, Yang, He, Hu, Zhang,
  Tan, Hu, and Yan]{yu2026duallatentmemoryvisual}
Xinlei Yu, Chengming Xu, Zhangquan Chen, Bo~Yin, Cheng Yang, Yongbo He, Yihao
  Hu, Jiangning Zhang, Cheng Tan, Xiaobin Hu, and Shuicheng Yan.
\newblock Dual latent memory for visual multi-agent system, 2026{\natexlab{a}}.
\newblock URL \url{https://arxiv.org/abs/2602.00471}.

\bibitem[Yu et~al.(2026{\natexlab{b}})Yu, Liu, Jin, Yuan, Kuang, and
  Wang]{yu2026learningcommunicateendtoendoptimization}
Ye~Yu, Heming Liu, Haibo Jin, Xiaopeng Yuan, Peng Kuang, and Haohan Wang.
\newblock Learning to communicate: Toward end-to-end optimization of
  multi-agent language systems, 2026{\natexlab{b}}.
\newblock URL \url{https://arxiv.org/abs/2604.21794}.

\bibitem[Yuan et~al.(2026)Yuan, Jin, Yu, Kuang, Yu, Dong, and
  Wang]{yuan2026closinglooplatentreasoning}
Xiaopeng Yuan, Haibo Jin, Ye~Yu, Peng Kuang, Lijun Yu, Yushun Dong, and Haohan
  Wang.
\newblock Closing the loop on latent reasoning via test-time reconstruction,
  2026.
\newblock URL \url{https://arxiv.org/abs/2606.06252}.

\bibitem[Zeng et~al.(2026)Zeng, Chen, Cheng, Zha, Liu, Chen, Yang, Dong, Zhang,
  and Tong]{zeng2026modelmindsunlockingmultiagent}
Zhichen Zeng, Huiyuan Chen, Jingru Cheng, Juan Zha, Ming Liu, Ying Chen, Xiyuan
  Yang, Chaosheng Dong, Haiyang Zhang, and Hanghang Tong.
\newblock One model, many minds: Unlocking multi-agent synergy in a single
  agent via mixture of roles, 2026.
\newblock URL \url{https://arxiv.org/abs/2608.27338}.

\bibitem[Zhai et~al.(2023)Zhai, Mustafa, Kolesnikov, and Beyer]{zhai2023siglip}
Xiaohua Zhai, Basil Mustafa, Alexander Kolesnikov, and Lucas Beyer.
\newblock Sigmoid loss for language image pre-training.
\newblock In \emph{Proceedings of the IEEE/CVF International Conference on
  Computer Vision (ICCV)}, pp.\  11975--11986, October 2023.
\newblock \doi{10.1109/ICCV51070.2023.01100}.

\bibitem[Zhang et~al.(2024{\natexlab{a}})Zhang, He, Qian, Li, Li, Qin, Kang,
  Ma, Liu, Lin, Rajmohan, Zhang, and Zhang]{zhang2024large}
Chaoyun Zhang, Shilin He, Jiaxu Qian, Bowen Li, Liqun Li, Si~Qin, Yu~Kang,
  Minghua Ma, Guyue Liu, Qingwei Lin, Saravan Rajmohan, Dongmei Zhang, and
  Qi~Zhang.
\newblock Large language model-brained gui agents: A survey.
\newblock \emph{arXiv preprint arXiv:2411.18279}, 2024{\natexlab{a}}.

\bibitem[Zhang et~al.(2025{\natexlab{a}})Zhang, Cui, Chen, Wang, Zhang, Wang,
  Wu, and Hu]{zhang2025stopovervaluingmultiagentdebate}
Hangfan Zhang, Zhiyao Cui, Jianhao Chen, Xinrun Wang, Qiaosheng Zhang, Zhen
  Wang, Dinghao Wu, and Shuyue Hu.
\newblock Stop overvaluing multi-agent debate -- we must rethink evaluation and
  embrace model heterogeneity, 2025{\natexlab{a}}.
\newblock URL \url{https://arxiv.org/abs/2502.08788}.

\bibitem[Zhang \& Emu(2026)Zhang and
  Emu]{zhang2026latentchannelsactuallycommunicate}
Huixiang Zhang and Mahzabeen Emu.
\newblock Do latent channels actually communicate? a causal audit of latent
  multi-agent llm, 2026.
\newblock URL \url{https://arxiv.org/abs/2607.26773}.

\bibitem[Zhang et~al.(2025{\natexlab{b}})Zhang, Zuo, He, Sun, Liu, Jiang, Fan,
  Tian, Jia, Li, Fu, Lv, Zhang, Zeng, Qu, Li, Wang, Wang, Long, Liu, Xu, Ma,
  Zhu, Hua, Liu, Li, Chen, Qu, Li, Chen, Yuan, Gao, Li, Ma, Cui, Liu, Qi, Ding,
  and Zhou]{zhang2025surveyrein}
Kaiyan Zhang, Yuxin Zuo, Bingxiang He, Youbang Sun, Runze Liu, Che Jiang,
  Yuchen Fan, Kai Tian, Guoli Jia, Pengfei Li, Yu~Fu, Xingtai Lv, Yuchen Zhang,
  Sihang Zeng, Shang Qu, Haozhan Li, Shijie Wang, Yuru Wang, Xinwei Long,
  Fangfu Liu, Xiang Xu, Jiaze Ma, Xuekai Zhu, Ermo Hua, Yihao Liu, Zonglin Li,
  Huayu Chen, Xiaoye Qu, Yafu Li, Weize Chen, Zhenzhao Yuan, Junqi Gao, Dong
  Li, Zhiyuan Ma, Ganqu Cui, Zhiyuan Liu, Biqing Qi, Ning Ding, and Bowen Zhou.
\newblock A survey of reinforcement learning for large reasoning models.
\newblock \emph{arXiv preprint arXiv:2509.08827}, 2025{\natexlab{b}}.

\bibitem[Zhang \& Math-AI(2025)Zhang and Math-AI]{aime25}
Yifan Zhang and Team Math-AI.
\newblock American invitational mathematics examination (aime) 2025.
\newblock \url{https://huggingface.co/datasets/math-ai/aime25}, 2025.
\newblock Dataset card on Hugging Face (license: Apache-2.0).

\bibitem[Zhang et~al.(2024{\natexlab{b}})Zhang, Sun, Chen, Pfister, Zhang, and
  Arik]{zhang2024chain}
Yusen Zhang, Ruoxi Sun, Yanfei Chen, Tomas Pfister, Rui Zhang, and Sercan Arik.
\newblock Chain of agents: Large language models collaborating on long-context
  tasks.
\newblock \emph{Advances in Neural Information Processing Systems},
  37:\penalty0 132208--132237, 2024{\natexlab{b}}.

\bibitem[Zhang et~al.(2025{\natexlab{c}})Zhang, Dai, Bo, Ma, Li, Chen, Zhu,
  Dong, and Wen]{zhang2025survey}
Zeyu Zhang, Quanyu Dai, Xiaohe Bo, Chen Ma, Rui Li, Xu~Chen, Jieming Zhu,
  Zhenhua Dong, and Ji-Rong Wen.
\newblock A survey on the memory mechanism of large language model-based
  agents.
\newblock \emph{ACM Transactions on Information Systems}, 43\penalty0
  (6):\penalty0 155:1--155:47, 2025{\natexlab{c}}.
\newblock \doi{10.1145/3748302}.

\bibitem[Zhang et~al.(2025{\natexlab{d}})Zhang, He, Yan, Shen, Zhao, Wang,
  Shen, and Wang]{zhang2025soft}
Zhen Zhang, Xuehai He, Weixiang Yan, Ao~Shen, Chenyang Zhao, Shuohang Wang,
  Yelong Shen, and Xin~Eric Wang.
\newblock Soft thinking: Unlocking the reasoning potential of llms in
  continuous concept space.
\newblock \emph{arXiv preprint arXiv:2505.15778}, 2025{\natexlab{d}}.

\bibitem[Zhao et~al.(2025{\natexlab{a}})Zhao, Foo, Hu, Theobalt, Rahmani, and
  Liu]{zhao2025llm}
Bingxi Zhao, Lin~Geng Foo, Ping Hu, Christian Theobalt, Hossein Rahmani, and
  Jun Liu.
\newblock Llm-based agentic reasoning frameworks: A survey from methods to
  scenarios.
\newblock \emph{arXiv preprint arXiv:2508.17692}, 2025{\natexlab{a}}.

\bibitem[Zhao et~al.(2025{\natexlab{b}})Zhao, Xie, Lei, Song, Shi, Li, Liu,
  Xie, and Zhang]{zhao2025connecting}
Jiaxing Zhao, Hongbin Xie, Yuzhen Lei, Xuan Song, Zhuoran Shi, Lianxin Li,
  Shuangxue Liu, Linguo Xie, and Haoran Zhang.
\newblock Cochain: Balancing insufficient and excessive collaboration in llm
  agent workflows.
\newblock \emph{arXiv preprint arXiv:2505.10936}, 2025{\natexlab{b}}.

\bibitem[Zhao et~al.(2025{\natexlab{c}})Zhao, Yuksekgonul, Wu, and
  Zou]{zhao2025sirius}
Wanjia Zhao, Mert Yuksekgonul, Shirley Wu, and James Zou.
\newblock Sirius: Self-improving multi-agent systems via bootstrapped
  reasoning.
\newblock \emph{arXiv preprint arXiv:2502.04780}, 2025{\natexlab{c}}.

\bibitem[Zheng et~al.(2025)Zheng, Zhao, Li, Xie, Gao, Zhang, and
  Zhang]{zheng2025thought}
Yujia Zheng, Zhuokai Zhao, Zijian Li, Yaqi Xie, Mingze Gao, Lizhu Zhang, and
  Kun Zhang.
\newblock Thought communication in multiagent collaboration.
\newblock \emph{arXiv preprint arXiv:2510.20733}, 2025.

\bibitem[Zhong et~al.(2024)Zhong, Guo, Gao, Ye, and Wang]{zhong2024memorybank}
Wanjun Zhong, Lianghong Guo, Qiqi Gao, He~Ye, and Yanlin Wang.
\newblock Memorybank: Enhancing large language models with long-term memory.
\newblock In \emph{Proceedings of the AAAI Conference on Artificial
  Intelligence}, volume~38, pp.\  19724--19731, 2024.

\bibitem[Zhou et~al.(2023)Zhou, Yan, Shlapentokh-Rothman, Wang, and
  Wang]{zhou2024languageagenttreesearch}
Andy Zhou, Kai Yan, Michal Shlapentokh-Rothman, Haohan Wang, and Yu-Xiong Wang.
\newblock Language agent tree search unifies reasoning acting and planning in
  language models, 2023.
\newblock URL \url{https://arxiv.org/abs/2310.04406}.

\bibitem[Zhou et~al.(2025)Zhou, Geng, Xue, Kang, Qin, Wang, Yin, and
  Bai]{zhou2025reso}
Heng Zhou, Hejia Geng, Xiangyuan Xue, Li~Kang, Yiran Qin, Zhiyong Wang, Zhenfei
  Yin, and Lei Bai.
\newblock Reso: A reward-driven self-organizing llm-based multi-agent system
  for reasoning tasks.
\newblock \emph{arXiv preprint arXiv:2503.02390}, 2025.

\bibitem[Zhou et~al.(2019)Zhou, Du, and Ren]{zhou2019improving}
Wenxuan Zhou, Junyi Du, and Xiang Ren.
\newblock Improving bert fine-tuning with embedding normalization.
\newblock \emph{arXiv preprint arXiv:1911.03918}, 2019.

\bibitem[Zhu et~al.(2025)Zhu, Hao, Hu, Jiao, Russell, and
  Tian]{zhu2025reasoning}
Hanlin Zhu, Shibo Hao, Zhiting Hu, Jiantao Jiao, Stuart Russell, and Yuandong
  Tian.
\newblock Reasoning by superposition: A theoretical perspective on chain of
  continuous thought.
\newblock \emph{arXiv preprint arXiv:2505.12514}, 2025.

\bibitem[Zhuang et~al.(2023)Zhuang, Chen, Yu, Mitra, Bursztyn, Rossi, Sarkhel,
  and Zhang]{zhuang2023toolchainefficientactionspace}
Yuchen Zhuang, Xiang Chen, Tong Yu, Saayan Mitra, Victor Bursztyn, Ryan~A.
  Rossi, Somdeb Sarkhel, and Chao Zhang.
\newblock Toolchain*: Efficient action space navigation in large language
  models with a* search, 2023.
\newblock URL \url{https://arxiv.org/abs/2310.13227}.

\bibitem[Zhuge et~al.(2024)Zhuge, Wang, Kirsch, Faccio, Khizbullin, and
  Schmidhuber]{zhuge2024language}
Mingchen Zhuge, Wenyi Wang, Louis Kirsch, Francesco Faccio, Dmitrii Khizbullin,
  and J{\"u}rgen Schmidhuber.
\newblock Language agents as optimizable graphs.
\newblock \emph{arXiv preprint arXiv:2402.16823}, 2024.

\bibitem[Zou et~al.(2023)Zou, Phan, Chen, Campbell, Guo, Ren, Pan, Yin,
  Mazeika, Dombrowski, Goel, Li, Byun, Wang, Mallen, Basart, Koyejo, Song,
  Fredrikson, Kolter, and Hendrycks]{zou2023representation}
Andy Zou, Long Phan, Sarah Chen, James Campbell, Phillip Guo, Richard Ren,
  Alexander Pan, Xuwang Yin, Mantas Mazeika, Ann-Kathrin Dombrowski, Shashwat
  Goel, Nathaniel Li, Michael~J. Byun, Zifan Wang, Alex Mallen, Steven Basart,
  Sanmi Koyejo, Dawn Song, Matt Fredrikson, J.~Zico Kolter, and Dan Hendrycks.
\newblock Representation engineering: A top-down approach to ai transparency.
\newblock \emph{arXiv preprint arXiv:2310.01405}, 2023.

\bibitem[Zou et~al.(2025)Zou, Yang, Qiu, Li, Tieu, Lu, Shen, Tong, Choi, He,
  Zou, Wang, and Yang]{zou2025latent}
Jiaru Zou, Xiyuan Yang, Ruizhong Qiu, Gaotang Li, Katherine Tieu, Pan Lu,
  Ke~Shen, Hanghang Tong, Yejin Choi, Jingrui He, James Zou, Mengdi Wang, and
  Ling Yang.
\newblock Latent collaboration in multi-agent systems.
\newblock \emph{arXiv preprint arXiv:2511.20639}, 2025.

\end{thebibliography}
\end{document}